\documentclass{article}

\usepackage{PRIMEarxiv}
\usepackage{adjustbox}
\usepackage{caption, subcaption}
\usepackage[section]{placeins}
\usepackage{natbib}
\usepackage[utf8]{inputenc} 
\usepackage[T1]{fontenc}    
\usepackage{hyperref}       
\usepackage{url}            
\usepackage{booktabs}       
\usepackage{amsfonts}       
\usepackage{nicefrac}       
\usepackage{microtype}      
\usepackage{lipsum}
\usepackage{fancyhdr}       
\usepackage{graphicx}       
\usepackage{amsmath}
\usepackage{amsthm}
\usepackage{wasysym}%
\usepackage{csquotes}
\MakeOuterQuote{"}
\usepackage{amsmath,amssymb,amsfonts}
\newtheorem{theorem}{Theorem}[section]
\newtheorem{example}[theorem]{Example}
\newtheorem{definition}[theorem]{Definition}

\hypersetup{
    colorlinks=false,
    pdfborder={0 0 0},
    pdfborderstyle={/S/U/W 0},
}

\title{Reranking individuals: The effect of fair classification within-groups}

\author{ \href{https://orcid.org/0000-0003-3784-826X}{\includegraphics[scale=0.06]{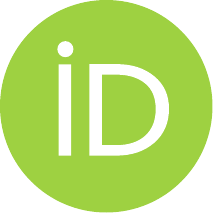}\hspace{1mm}Sofie~Goethals} \\
	University of Antwerp\\
	Antwerp, Belgium \\
	\texttt{sofie.goethals@uantwerpen.be} \\
        \And
        Marco Favier\\
	University of Antwerp\\
	Antwerp, Belgium \\
    \And
        Toon Calders\\
	University of Antwerp\\
	Antwerp, Belgium \\
	}

\begin{document}



\maketitle

\begin{abstract}
Artificial Intelligence (AI) finds widespread application across various domains, but it sparks concerns about fairness in its deployment. The prevailing discourse in classification often emphasizes outcome-based metrics comparing sensitive subgroups without consideration of the differential impacts \textit{within} subgroups. 
Bias mitigation techniques do not only affect the ranking of pairs of instances across sensitive groups, but often also significantly affect the ranking of instances within these groups. Such changes are hard to explain and raise concerns regarding the validity of the intervention. Unfortunately, these effects remain under the radar in the accuracy-fairness evaluation framework that is usually applied.
In this paper, we illustrate the effect of several popular bias mitigation methods, and how their output often does not reflect real-world scenarios.
\end{abstract}


\section{Introduction} \label{sec:intro}
In the rapidly evolving landscape of Artificial Intelligence (AI) and machine learning, the pursuit of fairness in algorithmic decision-making has emerged as a central concern. As the influence and scope of the decisions made by AI systems are increasing, there are growing concerns that the models making these decisions might unintentionally encode and even amplify human bias~\citep{corbett2018measure}.
\emph{Algorithmic bias} describes situations where sensitive groups are substantially disadvantaged by an algorithm or model. One of the ways bias can seep into a model is when it is trained on biased data, following the famous \emph{garbage in, garbage out} principle which emphasizing that flawed input data results in flawed output~\citep{geiger2020garbage}. Examples of biased AI models are everywhere, with cases in almost every domain. In the context of hiring, a well-known case is that of an automated Amazon recruitment system that had to be pulled because it was biased against female applicants~\cite{dastin2022amazon}. Much earlier already, St George’s Hospital Medical School’s Commission for Racial Equality discovered that a computer program used for initial screenings of applicants “written after careful analysis of the way in which the staff were making these choices” unfairly rejected women and individuals with non-European sounding names~\cite{johnson2021algorithmic, lowry1988blot}. There is an abundance of examples akin to these ones.

In this paper, we focus on \emph{fair classification}, which aims to ensure that algorithms make unbiased decisions across groups.
Many bias detection and mitigation methods exist, but most of them focus on "between-group fairness" where the main goal is to rectify disparities in model predictions between distinct demographic groups. This is undeniably critical, as it aims to rectify long-standing inequalities. However, it is also important to consider the complexities that exist "within" these groups, giving rise to a concept that is commonly referred to as "within-group fairness". \citet{speicher2018unified} already note that many approaches to group fairness tackle only between-group issues, worsening within-group fairness as a consequence. \citet{krco2023mitigating} highlight that the blind optimization of commonly used fairness metrics does not show who is impacted \textit{within} each group, while \citet{mittelstadt2023unfairness} emphasize that many of the currently used bias mitigation methods can make every group worse off. These issues illustrate why solely looking at fairness by measuring disparities between groups is not adequate.

While various benchmarking studies attempt to evaluate the performance of bias mitigation methods, they often fall short, comparing what can be described as `apples to oranges'.
This issue arises because different bias mitigation methods can significantly vary the number of positive instances, and by comparing them as-is, the actual situation faced by practitioners is not taken into account.
\citet{scantamburlo2024prediction} argue that the ultimate decision of an automated system is informed by the prediction model, but in nearly all cases is also influenced by additional parameters such as quota or business rules. They make a distinction between the prediction model and the decision-making system, and discuss how the field of fair machine learning tends to blur the boundaries between the two concepts~\citep{scantamburlo2024prediction}. 
\citet{kwegyir2023repairing} confirm that when deploying a classifier in the real world, practitioners typically need to tinker with the threshold to make sure the model predictions meet their domain-specific needs.

This is why directly comparing these methods on prediction labels as-is is not a good idea, yet this approach remains common in current  benchmarking studies~\citep{chen2023comprehensive, hort2022bia, hufthammer2020bias, reddy2022benchmarking, janssen2021bias, menon2018cost}.
The existing metrics fall short in acknowledging the inherent differences in the situations they compare, rendering them insufficient for determining the superior performance of one method over another. More importantly, they fail to take into account the constraints in positive decision rate that are faced in the real world. For example, our results  demonstrate that for the Adult Income dataset one bias mitigation method yields a positive decision rate of 0.5\% while another results in a positive decision rate of 39.3\%, whereas the actual positive rate of the dataset is 23.9\%. This illustrates that the process of enforcing fairness can significantly alter the overall selection rate of the model. If we would compare the accuracy of these two mitigation methods, we are actually examining two entirely distinct points of the ROC curve, both of which may be inapplicable in real-world settings.
Moreover, the majority of bias mitigation methods do not result in the satisfaction of the fairness metrics, necessitating additional post-hoc interventions after their application. To address these issues, one should separate the function of the prediction model from the decision-making context~\citep{scantamburlo2024prediction}. This is necessary to distinguish between performance changes that are solely due to the model selecting more or less individuals, and those caused by the model selecting `\textit{worse}' individuals.

Following this, we explore the distinct impact that each bias mitigation method has. These methods attempt to make the model output more fair by imposing constraints at the group level, but they also have a significant effect within each group. While the primary goal is to increase the selection of individuals from the protected group and reduce selections from privileged groups, these methods can also change which specific individuals are selected within each group.
\footnote{In some cases, due to changes in the global selection rate, fewer individuals from a protected group may be selected while still achieving greater overall fairness—an outcome that is far from ideal.}
This seemingly arbitrary reshuffling within each group, as a side effect of fair classification, is currently not studied. 
In this process, certain individuals (from both the protected and the privileged group) who were initially assigned a positive prediction label will now be labelled negatively, while some group members that initially received a negative label will be switched to a positive one.
This reranking process within each group is not per definition negative -as we will discuss- but it deserves more attention. 
Machine learning models not only return prediction labels, they also return prediction scores, which creates an inherent ranking of instances. We will compare the rankings generated by each bias mitigation method to analyze the effect within subgroups, independent of changes in the positive decision rate.

In summary, even though the primary purpose of using bias mitigation methods is typically to boost selection form protected group, these methods often yield unintended consequences.
In this paper, we discuss two of these consequences:
\begin{itemize}
    \item The global selection rate is often modified by the bias mitigation methods, which makes the resulting prediction labels not comparable and not applicable in real-world settings. We demonstrate this in Section~\ref{subsec:comparison}, and argue that no distinction is made between the prediction model and the decision-making context.
    \item The bias mitigation methods will not only change who is selected between the groups, but also who is selected within each group. We illustrate for several methods and datasets how some methods will shift the ranking within each group in Section~\ref{sec:res:who}. We discuss the theory behind this in Section~\ref{sec:theory}.
\end{itemize}

\section{Background}

Before we go further, it is important to define some of the key terminology that is often used in the fairness literature. A \textit{sensitive attribute} refers to a characteristic or feature of an individual that is considered sensitive, often with respect to potential discrimination. This can include attributes such as race, gender, age, religion, sexual orientation, or any other factor that could be the basis of unfair treatment. Consequently, a \textit{protected group} typically refers to the demographic group that is at risk of being unfairly treated or discriminated against based on their sensitive attribute, while the \textit{privileged group} is the demographic category that is typically not subject to unfair treatment based on that sensitive attribute. 
In this paper, we operate under the assumption of a single binary sensitive attribute, implying the existence of a protected group and a privileged group. 
\textit{Fairness metrics} are quantitative measures used to assess the fairness of AI models, while \textit{fairness (or bias) mitigation strategies} are techniques used to either learn an AI model that is fair by design or modify AI models to reduce bias.

In the computer science community, a plethora of fairness metrics have been proposed~\citep{corbett2018measure}. One of the most popular approaches is the \textit{group fairness metric}, which quantifies the fairness of a machine learning model across different demographic or sensitive groups, aiming to identify disparities in the outcome between these groups. 
One of the simplest and most commonly used definitions in this category is \emph{demographic parity} (or \emph{statistical parity}), which states that the positive decision rate must be the same regardless of the value of the protected attribute. In our example of hiring, this means that a model must invite equal percentages of white and black applicants for an interview (if race is the sensitive attribute) or of male and female applicants (if gender is the sensitive attribute).
Other commonly used metrics include \emph{equalized opportunity}, which states that there should be an equal proportion of true positives in both groups, and \emph{equalized odds}, which examines whether both the proportion of true positives and true negatives is equal across groups.
Besides these, many other fairness metrics exist, and the issue is that most of them cannot be optimized at the same time~\cite{kleinberg2016inherent}. Deciding upon a group fairness metric to optimize thus means already imposing a certain world view.
Another popular approach to assess fairness in machine learning models is \emph{individual fairness}, which demands that similar individuals receive similar outcomes in a decision-making process, regardless of their group membership~\citep{dwork2012fairness, binns2020apparent}. \citet{dwork2012fairness} argue that instead of focusing on a group, we tend to care more about the individual. This notion is related to the contributions of this paper, as we will also argue that satisfying group fairness metrics is not necessarily fair from the viewpoint of the individual. 

A common starting point for designing a fair algorithm is simply to exclude sensitive attributes from the model. However, the limitations of this approach have been commonly addressed, with the most fundamental limitation being \textit{the proxy problem}~\citep{prince2019proxy}. The proxy problem states that the omission of sensitive attributes can lead the machine learning model to rely on proxy variables that indirectly encode the information contained in the sensitive attribute and hence still introduce bias into the model's decision-making process. A classic example of the proxy problem is the use of zip codes in the United States as a proxy for racial information, as these two attributes tend to be heavily correlated.
This has prompted many to argue that proxies should be excluded from the dataset as well, however, this is very difficult to operationalize~\citep{corbett2018measure}. This is because every attribute used in the machine learning model can be at least partially correlated with the sensitive attribute; and often, even strongly correlated attributes may be considered legitimate factors on which to base decisions (for example, education in the case of hiring)~\citep{corbett2018measure}.

This illustrates that creating a fair machine learning model is a tedious process. In response, many bias mitigation methods that claim to improve fairness, have been introduced. We can divide most of them into three categories: preprocessing, inprocessing and postprocessing. Each category targets a different stage of the machine learning pipeline to ensure fairness.
The idea behind preprocessing methods is that they will change the representation of the data before the machine learning model is learned, and as such neutralize any prejudiced information that could affect the model's decision. Inprocessing methods improve fairness during the training process, by incorporating fairness constraints in the learning algorithm and striving for a balance between accuracy and fairness. Postprocessing methods intervene after the model has made its predictions, by adjusting the outcomes to satisfy fairness criteria. We will discuss the used bias mitigation methods in more detail in the Materials and Methods section. These bias mitigation methods focus on satisfying the aforementioned group fairness metrics that measure disparities between groups~\cite{chen2023comprehensive}.

Another research area that we should consider is the area of fair ranking.
\citet{yang2017measuring} measure whether a ranking is fair by comparing the distribution of protected and non-protected candidates on different prefixes of the list, while \citet{zehlike2017fa} illustrates how group fairness metrics can be satisfied for different prefixes of the ranking.
\citet{yang2019balanced} demonstrate how adding diversity constraints to ranking algorithms can reduce in-group fairness, a concept related to our measure of within-group fairness.
However, it is crucial to highlight the differences with this study, as we study fair classification, which focuses on equal outcomes between groups, and which does not entail that the ranking distribution should be fair.~\footnote{This can be fixed by postprocessing methods, for example by using different thresholds for each group, for the required positive decision rate. Fair ranking will be more strict, as it requires that the fairness metrics are satisfied for different prefixes of the ranked list. }  
We will show that several methods that are used for fair classification have as side effect that they change the rankings within each group, which in turn has influence on the final prediction labels of each individual. This seemingly arbitrary ‘reshuffling’ as a consequence of fair classification is currently not studied and deserves more attention.

 \section{Theory} \label{sec:theory}
 This section offers a theoretical foundation for analyzing how bias mitigation methods impact group-specific rankings.
 Such a framework highlights the importance of auditing fairness algorithms and explains why measuring differences in rankings provides meaningful insights into a model's behavior.
 The key takeaway from this section is that bias mitigation methods have two potential outcomes: they either address within-group bias or must align theoretically with Threshold Optimization to be optimal. A more comprehensive explanation of the theory underlying this section is available in \cite{favier2024patriarchy}.

Consider a space of individuals from whom we have collected attributes $X=(X_1,\dots, X_n)$, as well as an assumed-binary sensitive attribute $A$, and a binary label $Y\in\{0,1\}$.

Every individual has an unbiased and fair probability of receiving the positive label $p(Y=1\vert X=x,A=a)$. At the same time, we have access to a biased model, trained on biased data, that provides a score $S(x,a)\in [0,1]$, which can be used as a biased proxy for $p(Y=1\vert X=x, A=a)$.

More specifically, $S(x,a)$ is assumed to be the best fairness-agnostic model achievable, capable of providing the most accurate score based on biased data. In this sense, $S(x,a)$ accurately reflects the unfair probability distribution captured in the collected data. 
In other words, we will assume that $S(x,a)$ has no epistemic uncertainty, and any deviation of $S(x,a)$ from the fair $p(Y=1\vert X=x,A=a)$ is due to bias in the data.
Given a distribution on $X\times A$, both $p(Y=1\vert X=x, A=a)$ and $S(x,a)$ can be viewed as random variables on $[0,1]$, which we assume to be continuous.
\begin{example}[part 1]
We want to predict the income $(Y = \text{"high"}/\text{"low"})$ of individuals based on two independent attributes: their education, measured in years $(X \in \mathbb{N})$, and their gender $(A = \male / \female)$. With equal levels of education, individuals of different genders are, on average, equally capable and should therefore receive the same average income. However, due to societal biases, women earn, on average, 10\% less than their male counterparts at each education level.
Based on fair probabilities, we assert that in this example the following holds true:
\[
    p(Y=\text{"high"}\vert X, A=\male) = p(Y=\text{"high"}\vert X, A={\female})
\]
Notice that, since $X$ and $A$ are independent, fair probabilities satisfy demographic parity. However, if we were to train a model on biased data, we would have that
\[
    S(X, A={\female}) = 90\% \cdot S(X, A={\male}) 
\]
This occurs because the model has learned biased probabilities, and even with calibration on the data, it will remain biased. Furthermore, the scores no longer adhere to demographic parity.
\end{example}
\begin{definition} A \emph{decision} is a function $\hat Y\colon U\subseteq X\times A\to {0,1}$ that we can evaluate based on how well it performs on each label. In particular, for all $y\in\{0,1\}$, we define the following values:
\begin{align*}
    P_f(\hat{Y} = y\vert Y = y) &:= 
    \dfrac{\int_{U}\vert 1-y-\hat{Y}(x,a)\vert\cdot\vert 1-y-p(Y=1\vert X=x\vert A=a)\vert\text{d}(x,a)}
    {\int_{U}\vert 1-y-p(Y=1\vert X=x, A=a)\vert\text{d}(x,a)}
    \\
    P_u(\hat{Y} = y\vert Y = y) &:= 
    \dfrac{\int_{U}\vert 1-y-\hat{Y}(x,a)\vert\cdot\vert 1-y-S(x,a)\vert\text{d}(x,a)}{\int_{U}\vert 1-y-S(x,a)\vert\text{d}(x,a)}
\end{align*}
where $P_f(\hat{Y} = y\vert Y = y)$ is the true positive/negative rate according to fair probabilities, and $P_u(\hat{Y} = y\vert Y = y)$ is the true positive/negative rate according to the unfair score, depending on the considered label $y$. 
\end{definition}
This allows us to define the following:
\begin{definition}[Pareto Order]
    Given two decisions $\hat{Y},\hat{Y}'$, we will say that $\hat{Y}$ is Pareto \emph{non-inferior} to $\hat{Y}'$ according to fair probabilities, and write $\hat{Y}\geq_f \hat{Y}'$ if and only if 
    \[
        P_f(\hat{Y}= 1\vert Y=1)\geq P_f(\hat{Y}' = 1 \vert Y=1 )\text{ and }P_f(\hat{Y}= 0\vert Y=0)\geq P_f(\hat{Y}' = 0 \vert Y=0 )
    \]
Similarly, we will say that $\hat{Y}$ is Pareto \emph{non-inferior} to $\hat{Y}'$ according to unfair probabilities, and write $\hat{Y}\geq_u \hat{Y}'$ if and only if 
    \[
        P_u(\hat{Y}= 1\vert Y=1)\geq P_u(\hat{Y}' = 1 \vert Y=1 )\text{ and }P_u(\hat{Y}= 0\vert Y=0)\geq P_u(\hat{Y}' = 0 \vert Y=0 )
    \]
\end{definition}
It is possible to prove that Pareto maximal decisions according to fair probabilities are all and only those of the following form  almost everywhere:
\[
    \hat{Y}(x,a) = \begin{cases}
        1 & \text{if }p(Y=1\vert X=x, A=a) > \tau\\
        0 & \text{otherwise}
    \end{cases}
\]
for some $\tau\in [0,1]$. Similarly, for Pareto maximal decisions according to unfair probabilities, the same property holds with respect to $S(x,a)$.

\begin{example}[part 2]
    After training a model for the previous example, we obtain the following scores $S(x,a)$:
    \[
        S(x,a)=
        \begin{cases}
          x/50 & \text{if }a={\male}\\
          90\%x/50 & \text{if }a={\female}
        \end{cases}
    \]
    So the following decision function is used to make predictions:
    \[
      \hat{Y}(x,a) = \begin{cases}
        1 & \text{if }S(x,a)> 0.5\\
        0 & \text{otherwise}
      \end{cases}
    \]
    According to unfair probabilities, this decision is Pareto maximal. However, it is not maximal according to the fair probabilities, since according to fair probabilities the decision is equivalent to
    \[
        \hat{Y}(x,a) = \begin{cases}
            1 & \text{if }a=\male\text{ and }p(Y=1\vert X=x,A=a)> 0.5\\
            1 & \text{if }a=\female\text{ and }p(Y=1\vert X=x,A=a)> 0.55\overline{5}\\
            0 & \text{otherwise}
        \end{cases}
    \]
    which depends specifically on the sensitive attribute.
\end{example}

Under this framework, we can prove that the absence of within-group bias is equivalent to claiming that Pareto maximal decisions on fair probabilities correspond to decision that are still Pareto maximal according to unfair probabilities but only on each sensitive group. In other words, either we believe that within group bias exists or we already know the optimal approach to make fair decisions.

More formally, we formalize the assumption that every fair Pareto maximal decision corresponds to unfair Pareto maximal ones for each sensitive group as follows
\begin{definition}[Affirmative Action Assumption]
  Let $\hat Y\colon X\times A\to \{0,1\}$ be a maximal decision according to fair probabilities. For every $a\in A$, there exists a maximal decision according to unfair probabilities $\hat{Y}_{a}\colon X\times \{a\}\to \{0,1\}$ such that 
  $
    \hat{Y}= \bigcup_{a\in A}\hat{Y}_a
  $
\end{definition}
\begin{example}[part 3]
  In our running example we have that the Affirmative Action Assumption does hold, since the following holds:
  \[
    \hat{Y}(x,a) = \begin{cases}
      1 & \text{if }p(Y=1\vert X=x, A=a) > \tau\\
      0 & \text{otherwise}
  \end{cases}
  \iff
  \hat{Y}(x,a) = \begin{cases} 1 & \text{if }a=\male\wedge S(x, a) > \tau\\
    1 & \text{if }a=\female\wedge S(x, a) > 9\tau/10\\
    0 & \text{otherwise}
\end{cases}
  \]
which means that each fair optimal decision is equivalent to a decision that utilizes two different thresholds, one for each sensitive group.
\end{example}
The following theorem asserts that if we have a Pareto maximal decision according to fair probabilities, then, in absence of within group bias, this decision can be decomposed into two maximal decisions - one for each sensitive group.
\begin{theorem}\label{teo:condition}
	The Affirmative Action assumption holds if and only if for all $a\in A$ it holds that for almost all $(x,a), (x',a)\in X\times \{a\}$
  \[
    p(Y=1\vert X=x, A=a)\leq p(Y=1\vert X=x', A=a)\Rightarrow S(x,a)\leq S(x',a)
  \] 
  \begin{proof}
      The Proof of this theorem can be found in Section~\ref{sec:app:proof}.
  \end{proof}
\end{theorem}



This Theorem implies that reranking within each group only makes sense when the Affirmative Action assumption does not hold, and there is some form of within-group bias. When this is the case, it makes sense that we need to make changes within each group as well. However, when the Affirmative Action assumption holds and there is no within-group bias, there is no practical reason to use any fairness intervention other than Threshold Optimization, as research has shown that for fairness measures such as Demographic Parity and Equal Opportunity, Threshold Optimization can always find an optimal solution.
We will see in Section~\ref{sec:res:who} that many of the used bias mitigation methods do lead to reranking of individuals within each group, while they do not explicitly assume any within-group bias.
\section{Materials and methods}
\subsection{Materials}

We will use several real world datasets that are common in the domain of fair machine learning~\citep{le2022survey}.
The \textbf{Adult Income} dataset contains information extracted from the 1994 census data  with as target variable whether the income of a person exceeds \$50,000  a year or not. 
The \textbf{Compas} dataset includes demographic information
and criminal histories of defendants from Broward County, and is used to predict whether a
defendant will re-offend within two years.
The \textbf{Dutch Census} dataset represents aggregated groups of people in the Netherlands for the year 2001, and can be used to predict whether a person's occupation can be categorized as a high-level (prestigious) or a low-level profession~\citep{van20002001}.
The \textbf{Law admission }dataset contains a Law School Admission Council (LSAC) survey conducted across 163 law schools in the United States in 1991~\citep{wightman1998lsac} and can be used to predict whether the student will pass the bar exam or not.
The \textbf{Student Performance} dataset describes the achievements of students in two Portuguese schools~\citep{cortez2008using}. The classification task is to predict whether they score above average in mathematics.

\begin{table*}[ht]
        \caption{Used datasets}
    \label{tab:my_label}
     \centering
    \begin{adjustbox}{width=\linewidth}
    \begin{tabular}{l|llllll}
    \hline
         Name&  \# instances&  \# attributes &Protected attribute  &Protected group&  Target attribute & Base rate\\ \hline
         Adult&  48,842&  10 &Gender  &Female&  High income & 23.93\%\\ 
         Compas&  5,278&   7&Race  &African-American&  Low risk & 52.16\%\\
         Dutch Census&  60,420&   11&Gender  &Female&  High occupation & 52.39\%\\
         Law admission&  20,798&  11 &Race &Non-White&  Pass the bar & 88.97\%\\ 
         Student Performance & 649 & 29 & Gender & Male & High score in mathematics & 53.62\% \\ \hline
    \end{tabular}
    \end{adjustbox}
\end{table*}

\subsection{Methods}
\subsubsection{Machine learning models}
We train fully connected neural networks on each dataset, utilizing binary cross-entropy as the loss metric.\footnote{We use a neural network, with one hidden layer with 200 nodes. We use 50 epochs, a batch size of 128 and the Adam optimizer. We do not perform hyperparameter tuning.} 
Neural networks are chosen because the implementation of one the bias mitigation methods (Adversarial Debiasing) requires this.
Although the remaining methods are model-agnostic, we opt for consistency in our approach, employing a neural network across all bias mitigation methods to ensure comparability in our final results.

\subsubsection{Bias mitigation methods} \label{subsec:bias_mit_methods}
Numerous debiasing strategies currently exist, but we focus on the methods available in the AIF360 package~\citep{bellamy2018ai}. For all algorithms, we use the default settings.

 As \textbf{preprocessing} methods, we use \emph{Learning Fair Representations} (\textbf{LFR}) and \emph{Disparate Impact Remover} (\textbf{DIR}). The idea behind Learning Fair Representations~\citep{zemel2013learning} is that a new representation Z is learned that removes the information correlated with the sensitive attribute, but preserves the other information about X as much as possible. Disparate Impact Remover~\citep{feldman2015certifying} modifies the training data to reduce the influence of sensitive attributes, but preserves rank-ordering within groups.
 
 We use the \textbf{inprocessing} methods of \textit{Adversarial Debiasing} (\textbf{ADV}) and the \textit{Meta Fair Classifier} (\textbf{MFC}).
 Adversarial Debiasing~\citep{zhang2018mitigating} combines a classifier that predicts the class label with an adversary that predicts the sensitive attribute. The goal is to maximize the classifier's performance while minimizing that of the adversary. The Meta Fair Classifier~\citep{celis2019classification} takes the fairness metric as part of the input and returns a classifier optimized with respect to that fairness metric.

 As \textbf{postprocessing} methods, we use \textit{Equalized Odds Postprocessing} (\textbf{EOP}), \textit{Reject Option Classification} (\textbf{ROC}) and \textit{Threshold Optimation} (\textbf{TO}). Equalized Odds Postprocessing~\citep{hardt2016equality} will solve a linear program to find the probabilities with which to change output labels in order to optimize equalized odds, while Reject Option Classification~\citep{kamiran2012decision} will flip the predictions the model is not confident of.\footnote{Reject option classification identifies instances where the model is uncertain about its prediction and essentially `rejects' making a definite decision. In this implementation, aiming to enhance fairness, the labels of these instances are flipped to satisfy a fairness criteria.}
Threshold Optimization~\citep{kamiran2012decision} is maybe the most straightforward method of mitigating bias as it will optimizes the thresholds for both groups in isolation.\footnote{For this bias mitigation method, we use the implementation available through the Fairlearn toolbox~\cite{bird2020fairlearn} as this method is not available in the AIF360 toolbox.} ROC and TO are implemented to enforce Demographic Parity, while EOP enforces Equalized Odds by default.

\subsection{Metrics} \label{subsec:metrics}
\subsubsection{Performance metrics}
Most benchmark studies compare the different mitigation methods on accuracy, which measures how often the prediction label assigned by the machine learning model coincides with the true label~\citep{chen2023comprehensive,hort2021fairea,krco2023mitigating}. 
It is commonly acknowledged that accuracy is not always the best metric to measure the performance of a machine learning model, as it is for example not suitable to deal with imbalanced class distributions (as in this case a model can obtain a high accuracy by just predicting all samples as the majority class)~\citep{chen2023comprehensive, mittelstadt2023unfairness}. This has led some studies to include other metrics such as the F1-score, Precision, or Recall~\citep{chen2023comprehensive}.
However, another notable drawback of these performance metrics is that they measure the performance at a specific classification threshold, as they use the prediction labels of the machine learning model and not the prediction scores. We can evaluate the performance of the prediction scores by using the Area Under the ROC Curve (AUC). \footnote{Note that this ROC, which stands for Receiver Operating Characteristic is different from the Reject Option Classification, that we also shorten as ROC.}

AUC allows for an objective comparison across classifiers, as it is unaffected by the choice of threshold or the frequency of classes~\citep{hand2009measuring}. It measures how well the prediction scores (and thus the ranking) of a machine learning model distinguishes between positive and negative cases. The formula for the \textbf{AUC score}, where $S(x_{i},a_i)$ notes the prediction score of instance $i$ where $x_i$ are the attributes of the individual and $a_i$ is their sensitive attribute: \\
\begin{equation*}
        P(S(x_{i}) > S(x_{z})| y_{i}=1, y_{z}=0)
\end{equation*}
This formula means that the AUC score is equivalent to the probability that a classifier will rank a randomly chosen positive instance higher than a randomly chosen negative instance.
\citet{provost1998case} previously advocated for the adoption of AUC as a standard for comparing classifiers in the broader field of machine learning. Despite this, its integration into Fair Machine Learning has been limited.\footnote{One example is where \citet{fong2021fairness} investigate how acquiring additional features can improve the AUC of the disadvantaged group.} Furthermore, the specific context of Fair Machine Learning presents additional justifications for focusing on the prediction scores generated by machine learning models, rather than solely on their prediction labels, due to unrealistic positive rates and the unsatisfaction of required fairness metrics.
In this study, we operate under the assumption that label bias is absent, meaning that the actual labels accurately represent the intended prediction target~\citep{wick2019unlocking,favier2023fair, lenders2023real}. Note that if label bias was present, it would compromise the validity of both the AUC and accuracy metrics, as these measures rely on these labels for their calculations.

Lastly, we will measure the positive decision rate (or positive classification rate) on the whole population. As mentioned earlier, this is important to be able to compare the different bias mitigation methods. We use $Y$ to denote the actual target label, and $\hat{Y}$ to denote the predicted label by the machine learning model. $A$ represents the sensitive attribute, where $a$ represents the protected group and $\neg a$ the privileged group.
The formula for the \textbf{Positive Decision Rate (PDR)}:
\begin{equation*}
        P(\hat{Y} = 1)
\end{equation*}

\subsubsection{Fairness metrics}
Many possible fairness metrics exist, but we will report two metrics that are commonly used in the fairness domain to measure disparities between groups:
Demographic parity (or statistical parity) states that the positive decision rate must be approximately the same in the protected group as in the privileged group. Statistical Parity Difference measures the difference in this positive decision rate between both groups.
The formula for the \textbf{Statistical Parity Difference } (SPD):\\
\begin{equation*}
        P(\hat{Y} = 1 | A = a) - P(\hat{Y} = 1 | A = \neg a)
\end{equation*}
Equalized opportunity requires the true positive rate to be approximately the same across groups.
The formula for \textbf{Equal Opportunity Difference} (EOD):\\
\begin{equation*}
        P(\hat{Y} = 1 | A = a, Y = 1) - P(\hat{Y} = 1 | A = \neg a, Y = 1)
\end{equation*}
Larger values of these metrics correspond to a higher level of bias towards one of the sensitive groups~\citep{hort2021fairea}.

\subsubsection{Rank correlation}
We can measure the correlation between two ranked lists by using the Kendall Tau metric, denoted as $\tau$. This metric measures the similarity between the orderings of two lists by quantifying the number of pairwise disagreements between them.

The formula for the \textbf{Kendall Tau coefficient} ($\tau$) is defined as:
\begin{equation*}
\frac{\text{number of concordant pairs} - \text{number of discordant pairs}}{\text{number of pairs}}
\end{equation*}

A concordant pair refers to a pair of observations where the order of the ranks is the same between both lists, while a discordant pair refers to a pair of observations where the order of the ranks is reversed between the two lists.
A $\tau$ value of 1 indicates perfect agreement between two rankings, whereas a value of -1 indicates perfect disagreement. Values closer to zero suggest little to no correlation, implying a lack of consistency in the ranking order between the two groups being compared.

\section{Results} \label{sec:Results} 
\subsection{Whom do the bias mitigation strategies affect?} \label{sec:res:who} 
The lack of transparency in the field of fair machine learning has been acknowledged in literature~\citep{rudin2020age, wachter2021fairness,goethals2023precof}. Other studies have already specifically criticized the opacity in the effects of the different bias mitigation methods~\citep{krco2023mitigating,favier2023fair, holstein2019improving}. What is the impact of each method, not only between sensitive groups but also within each group?

We provide transparency into the operational dynamics of the different bias mitigation methods, by comparing the score distributions after deploying each bias mitigation method with the score distributions of the initial machine learning model.
We illustrate the prediction scores from the initial ML model on the x-axis and the scores post-application of various bias mitigation methods on the y-axis. Additionally, we categorize the instances based on their affiliation with either the protected group (represented in dark blue) or the privileged group (represented in light blue). The chart is divided into four quadrants, each depicting the classification of instances as either positive or negative by the initial ML model and by each bias mitigation method.\footnote{Note that we use the custom threshold of $0.5$ here to generate the labels, based on the prediction scores. However, as argued by \citet{scantamburlo2024prediction}, this does not take into account the real constraints of  decision-making context.}  We also include a diagonal line that would contain all the instances, if the prediction scores would remain identical.
The presented results in Figure~\ref{fig:scores_compas}
 are calculated using the Compas dataset. The figures for the other datasets can be found in the Appendix (Figures~\ref{fig:scores_adult}, \ref{fig:scores_dutch}, \ref{fig:scores_law} and \ref{fig:scores_student}), with the results being consistent to the results of the Compas dataset.

\subsubsection*{Insights into the operational dynamics of each bias mitigation method}
\begin{figure*} [ht]
  \centering
    \begin{subfigure}{0.32\linewidth}
    \centering
    \includegraphics[width=\linewidth]{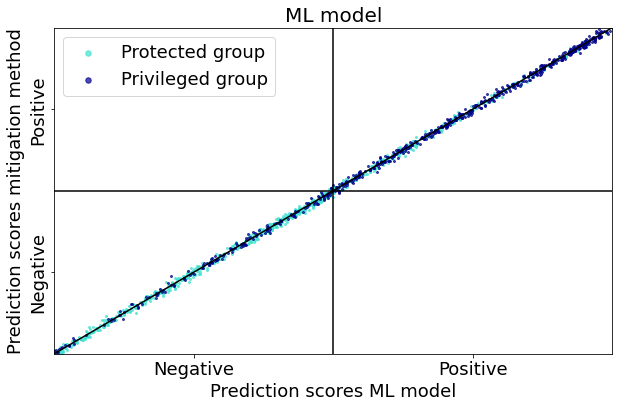}
    \caption{ML model}
    \label{subfig:ml}
  \end{subfigure}
  \begin{subfigure}{0.32\linewidth}
    \centering
    \includegraphics[width=\linewidth]{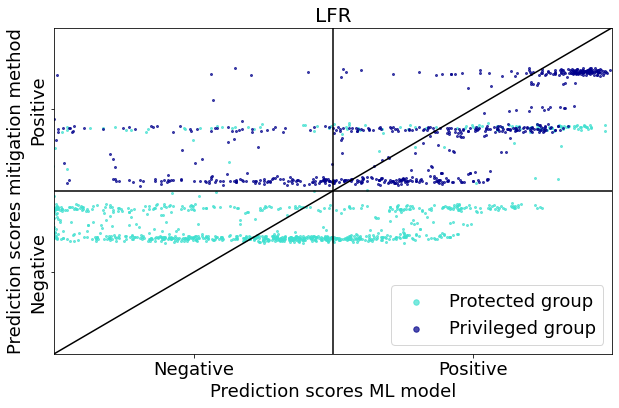}
    \caption{Preprocessing}
    \label{subfig:lfr}
  \end{subfigure}
    \begin{subfigure}{0.32\linewidth}
    \centering
    \includegraphics[width=\linewidth]{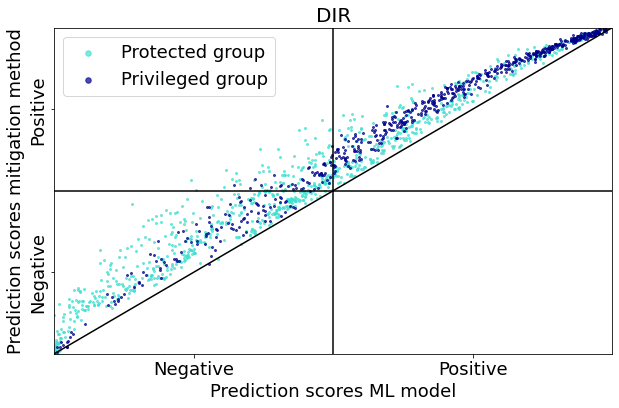}
    \caption{Preprocessing}
    \label{subfig:dir}
  \end{subfigure}
    \begin{subfigure}{0.32\linewidth}
    \centering
    \includegraphics[width=\linewidth]{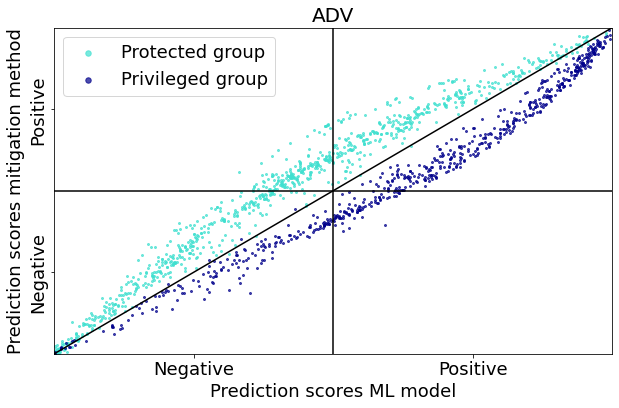}
    \caption{Inprocessing}
    \label{subfig:adv}
  \end{subfigure}
  \begin{subfigure}{0.32\linewidth}
    \centering
    \includegraphics[width=\linewidth]{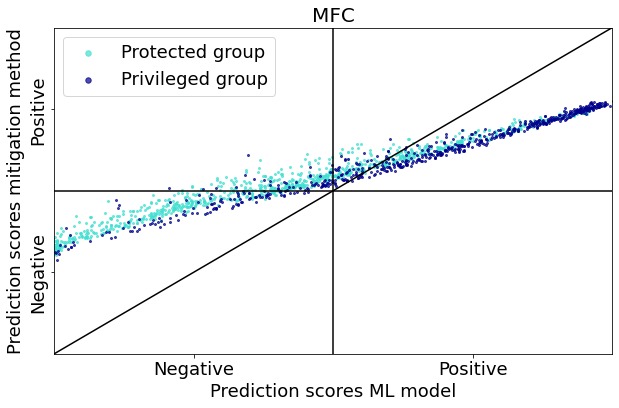}
    \caption{Inprocessing}
    \label{subfig:mfc}
  \end{subfigure} \\
    \begin{subfigure}{0.32\linewidth}
    \centering
    \includegraphics[width=\linewidth]{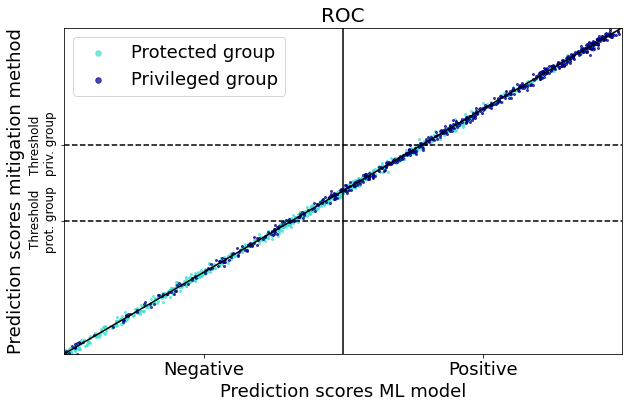}
    \caption{Postprocessing}
    \label{subfig:roc}
  \end{subfigure}
      \begin{subfigure}{0.32\linewidth}
    \centering
    \includegraphics[width=\linewidth]{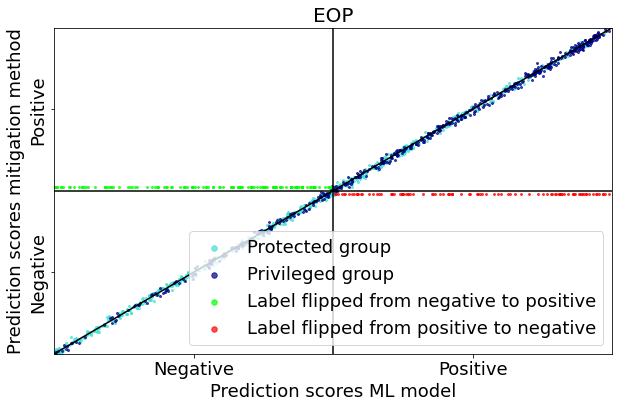}
    \caption{Postprocessing}
    \label{subfig:eop}
  \end{subfigure}
  \begin{subfigure}{0.32\linewidth}
    \centering
    \includegraphics[width=\linewidth]{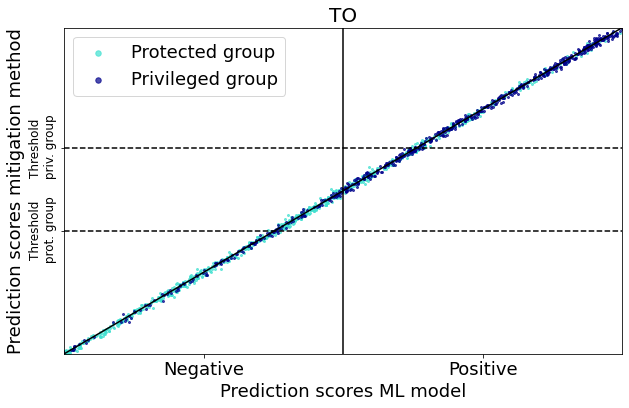}
    \caption{Postprocessing}
    \label{subfig:to}
  \end{subfigure}
  \caption{Score distributions for the Compas dataset. The x-axis represents the prediction scores of the initial ML model, while the y-axis represents the prediction score after applying each bias mitigation method. The second quadrant represents the instances that are `upgraded' by the bias mitigation method (initially predicted as negative, and after using the bias mitigation method predicted as positive), while the fourth quadrant represents the instances that are `downgraded' by the bias mitigation method (initially predicted as positive, and after the bias mitigation method predicted as negative).} 
  \label{fig:scores_compas}
  \end{figure*}

In Figure~\ref{subfig:ml}, we observe the prediction scores of the initial ML model.
By default, machine learning models use a classification threshold of $0.5$, categorizing instances above this threshold as positive and those below it as negative.\footnote{Note that this is not how it will actually be implemented in deployment. The notion of a decision threshold is not part of the model but is part of the decision logic. This is why it does not make sense to compare bias mitigation methods based on prediction labels generated by the machine learning model, without taking into account the decision-making context.} 

In Figures~\ref{subfig:lfr}-\ref{subfig:mfc}, we can assess the relation between the prediction scores of the initial ML model and the prediction scores after applying preprocessing (LFR and DIR) or inprocessing (ADV and MFC) bias mitigation methods.
Figure~\ref{subfig:lfr} reveals that LFR significantly alters the prediction scores, in a seemingly random way. Instances that receive a positive label post-LFR are notably different from those in the initial score ranking. This suggests a substantial transformation in the feature space due to the preprocessing method.  We will assess how this affects the performance and fairness metrics in Table~\ref{tab:results}. 
As shown in Figures~\ref{subfig:dir}, \ref{subfig:adv}, and \ref{subfig:mfc}, the methods DIR, ADV, and MFC also introduce significant alterations to the prediction scores, and thus the inherent ranking of instances. However, the altered prediction scores align more closely with the initial prediction scores compared to those obtained using LFR.

Figure~\ref{fig:scores_compas} also provides us with insights into the fundamental distinction among various bias mitigation methods—specifically, whether they impact the prediction scores or only the prediction labels.
Unlike preprocessing and inprocessing methods, the used postprocessing methods do not alter the inherent ranking; instead, they adjust labels based on the prediction scores of the initial machine learning model.
Each method implements a unique strategy for label flipping, all with the objective of meeting specified fairness metrics. For instance, ROC (Figure~\ref{subfig:roc}) targets the most uncertain instances for label flipping, TO (Figure~\ref{subfig:to}) establishes distinct thresholds for each group, while EOP (Figure~\ref{subfig:eop}) determines the ideal quantity of labels to flip within each group and executes these flips randomly.
In the scenarios of ROC and TO, group-specific new thresholds are determined (as illustrated in Figures ~\ref{subfig:roc} and \ref{subfig:to}). Regarding EOP, label flipping occurs randomly among negatively classified individuals in the protected group and positively classified individuals in the privileged group (as shown in green and red in Figure~\ref{subfig:eop}). In this case, the initial ranking impacts the initial labels, but will have no influence on the decision regarding which labels are flipped.\footnote{Note that with this postprocessing methods, essentially some reranking does happen, as individuals with a lower prediction score can now receive a positive label, while individuals from the same group with a higher prediction score do not. However, it is not reflected in the prediction scores, so the ranking similarity will be 1.}

Our examination provides insights into the specific instances affected by each bias mitigation method, in response to the work of~\citet{krco2023mitigating}. Take, for instance, the case of ROC: the instances that are flipped from a negative classification to a positive classification, are those instances that were initially classified as negative but that posses the highest prediction score according to the initial machine learning model. 
We also show in Figure~\ref{fig:scores_compas} that reranking does happen for some of the methods, but in the next section, we quantify how much. 

\subsubsection*{How similar are the score distributions?}

Instead of visualizing how the prediction scores of the initial ML relate to the prediction scores after each bias mitigation method, we can also calculate the overlap between the two rankings by utilizing the Kendall-Tau statistic~\citep{kendall1948rank}. 
The outcomes are presented in Table~\ref{tab:rbo_ranking}.
 Table~\ref{tab:rbo_ranking} illustrates that the ranking produced by LFR consistently presents the least similarity with the ranking produced by the initial machine learning model. In contrast, the use of DIR leads to a significantly higher degree of overlap in rankings across all datasets. The similarity in rankings when applying ADV and MFC falls into a more moderate category. These observations highlight the varying degrees of impact each method exerts on the prediction scores and thus the ranking of individuals.
 With postprocessing methods, the ranking remains identical to the ranking of the initial model, resulting in a ranking similarity of 1 (refer to Table~\ref{tab:rbo_ranking}), and the AUC remains consistent with the initial  model (see Table~\ref{tab:results}).
Our analysis reveals that across all datasets, the prediction scores and consequently the intrinsic rankings are substantially modified by every preprocessing and inprocessing bias mitigation method we evaluated.\footnote{We see that in one case (MFC for the Law dataset), the scores are even completely shuffled in the opposite way, which shows how random some of the effects can be.}
 While such modifications are not necessarily worrisome, it is crucial to assess whether they enhance or worsen the ranking, and to what extent. It also does not feel fair from the viewpoint of the individual that they are suddenly downweighted by a random intervention. \footnote{So far, we emphasized that only preprocessing and inprocessing methods suffer from this reranking process. However, the postprocessing method EOP also significantly suffers from this arbitrariness, as it will execute random flips within each group until a fairness metric is satisfied. This means that the impacted individuals can be different in each run, and can be individuals with a very high prediction score can be `downweighted' and individuals with a very low prediction score can be `uplifted' (as the prediction scores are not taken into account to determine who should be flipped). }
 Table~\ref{tab:rbo_ranking} displays the correlation between each of the bias mitigation methods and the prediction scores of the machine learning model, but we also visualise the correlation between the bias mitigation methods themselves in Figure~\ref{fig:correlation_plots}.

\begin{table*}[ht]
\centering
\caption{Table with the similarity between the initial ranking produced by the machine learning model and the ranking produced by the model after using a bias mitigation method, measured by the Kendall-Tau statistic. The values between brackets present the similarity in ranking for the protected group and privileged group respectively. }
\label{tab:rbo_ranking}
\begin{tabular}{@{}l|lllll@{}}
\toprule
Dataset & LFR                                                            & DIR                                                            & ADV                                                            & MFC                                                            & ROC-EOP-TO \\ \midrule
Adult   & \begin{tabular}[c]{@{}l@{}}0.318\\ (0.002, 0.034)\end{tabular} & \begin{tabular}[c]{@{}l@{}}0.770\\ (0.726, 0.761)\end{tabular} & \begin{tabular}[c]{@{}l@{}}0.729\\ (0.780,0.812)\end{tabular}  & \begin{tabular}[c]{@{}l@{}}0.638\\ (0.640, 0.663)\end{tabular} & 1          \\ \hline
Compas  & \begin{tabular}[c]{@{}l@{}}0.408\\ (0.153,0.534)\end{tabular}  & \begin{tabular}[c]{@{}l@{}}0.917\\ (0.883, 0.960)\end{tabular} & \begin{tabular}[c]{@{}l@{}}0.805\\ (0.926, 0.946)\end{tabular} & \begin{tabular}[c]{@{}l@{}}0.909\\ (0.900, 0.935)\end{tabular} & 1          \\ \hline
Dutch   & \begin{tabular}[c]{@{}l@{}}0.361\\ (-0.007, 0-0.004)\end{tabular} & \begin{tabular}[c]{@{}l@{}}0.992\\ (0.991, 0.991)\end{tabular} & \begin{tabular}[c]{@{}l@{}}0.808\\ (0.812, 0.913)\end{tabular} & \begin{tabular}[c]{@{}l@{}}0.819\\ (0.850, 0.786)\end{tabular} & 1          \\ \hline
Law     & \begin{tabular}[c]{@{}l@{}}0.570\\ (0.622, 0.523)\end{tabular} & \begin{tabular}[c]{@{}l@{}}0.870\\ (0.882, 0.858)\end{tabular} & \begin{tabular}[c]{@{}l@{}}0.908\\ (0.896, 0.907)\end{tabular} & \begin{tabular}[c]{@{}l@{}}-0.758\\ (-0.872, -0.737)\end{tabular} & 1          \\ \hline
Student & \begin{tabular}[c]{@{}l@{}}0.215\\ (0.248, 0.216)\end{tabular} & \begin{tabular}[c]{@{}l@{}}0.890\\ (0.891, 0.891)\end{tabular} & \begin{tabular}[c]{@{}l@{}}0.515\\ (0.731, 0.678)\end{tabular} & \begin{tabular}[c]{@{}l@{}}0.775\\ (0.843, 0.794)\end{tabular} & 1          \\ \bottomrule
\end{tabular}
\end{table*}

\subsection{Can we compare the bias mitigation methods?} \label{subsec:comparison}

\begin{table*}[ht]
\centering
\caption{Results of the bias mitigation strategies on the five datasets. We report: the AUC score overall, and split over the protected group and the privileged group, the accuracy (ACC), statical parity difference (SPD), equal opportunity difference (EOD) and the positive decision rate (PDR). Best values are highlighted in bold.}
\label{tab:results}
    \centering
    \begin{adjustbox}{width=\linewidth}
    \begin{tabular}{l|lllllllll}
\hline
   Dataset &                  Metric &   ML model &          LFR &          DIR &          ADV &          MFC &          ROC &          EOP &           TO \\
\hline \hline
 Adult     & AUC                     & 0.843          & 0.623        & \textbf{0.85 }        & 0.847        & 0.826        & 0.843        & 0.843        & 0.843        \\
           & $AUC^{pro} , AUC^{pri}$ & 0.811, 0.826   & 0.501, 0.508 & 0.824, 0.835 & \textbf{0.834, 0.843} & 0.789, 0.816 & 0.811, 0.826 & 0.811, 0.826 & 0.811, 0.826 \\
           & ACC                     & 0.806          & 0.766        & \textbf{0.825 }       & 0.82         & 0.81         & 0.728        & 0.77         & 0.793        \\
           & SPD                     & -0.26          & \textbf{-0.005  }     & -0.164       & -0.09        & -0.139       & -0.055       & -0.046       & \textbf{-0.005  }     \\
           & EOD                     & -0.139         & \textbf{0.0 }         & -0.181       & -0.274       & -0.278       & -0.332       & -0.283       & -0.368       \\
           & PDR                      & 0.247          & 0.005        & 0.174        & 0.161        & 0.178        & 0.393        & 0.151        & 0.167        \\ \hline
 Compas    & AUC                     & \textbf{0.834 }         & 0.693        & 0.833        & 0.808        & 0.832        & 0.834        & 0.834        & 0.834        \\
           & $AUC^{pro} , AUC^{pri}$ & \textbf{0.814}, 0.821   & 0.588, 0.703 & 0.81, \textbf{0.823 } & 0.809, 0.815 & 0.812, 0.821 & 0.814, 0.821 & 0.814, 0.821 & 0.814, 0.821 \\
           & ACC                     & \textbf{0.758 }         & 0.645        & 0.753        & 0.737        & 0.736        & 0.732        & 0.683        & 0.727        \\
           & SPD                     & -0.376         & -0.866       & -0.332       & -0.064       & -0.211       & -0.031       & -0.07        &\textbf{ 0.014  }      \\
           & EOD                     & -0.13          & \textbf{-0.055}       & -0.131       & -0.231       & -0.175       & -0.233       & -0.25        & -0.262       \\
           & PDR                      & 0.518          & 0.49         & 0.617        & 0.58         & 0.696        & 0.528        & 0.545        & 0.569        \\ \hline
 Dutch     & AUC                     & \textbf{0.887 }         & 0.657        & \textbf{0.887 }       & 0.874        & 0.883        & 0.887        & 0.887        & 0.887        \\
           & $AUC^{pro} , AUC^{pri}$ & 0.884, 0.848   & 0.499, 0.498 & 0.884, 0.849 & 0.881, 0.847 & \textbf{0.89, 0.852}  & 0.884, 0.848 & 0.884, 0.848 & 0.884, 0.848 \\
           & ACC                     & \textbf{0.812 }         & 0.476        & 0.786        & 0.768        & 0.695        & 0.776        & 0.754        & 0.763        \\
           & SPD                     & -0.318         & \textbf{0.0 }         & -0.432       & -0.171       & -0.394       & -0.066       & -0.159       & -0.02        \\
           & EOD                     & -0.026         & -0.315       & -0.047       & -0.073       & \textbf{-0.024 }      & -0.217       & -0.243       & -0.254       \\
           & PDR                      & 0.416          & 1.0          & 0.586        & 0.303        & 0.203        & 0.395        & 0.45         & 0.396        \\ \hline
 Law       & AUC                     & 0.882          & 0.83         & \textbf{0.883 }       & 0.879        & 0.122        & 0.882        & 0.882        & 0.882        \\
           & $AUC^{pro} , AUC^{pri}$ & 0.848, \textbf{0.864 }  & 0.803, 0.792 & \textbf{0.853, 0.864} & 0.843, 0.862 & 0.146, 0.142 & 0.848, 0.864 & 0.848, 0.864 & 0.848, 0.864 \\
           & ACC                     & \textbf{0.903 }         & 0.897        & \textbf{0.903 }       & 0.901        & 0.22         & 0.772        & 0.879        & 0.892        \\
           & SPD                     & -0.197         & -0.207       & -0.184       & -0.141       & 0.494        & -0.044       & -0.021       & \textbf{-0.002}       \\
           & EOD                     & -0.111         & -0.128       & -0.122       & -0.143       & -0.151       & \textbf{-0.104 }      & -0.197       & -0.21        \\
           & PDR                      & 0.954          & 0.967        & 0.961        & 0.965        & 0.289        & 0.711        & 0.961        & 0.979        \\ \hline
Student   & AUC                     & 0.803          & 0.693        & \textbf{0.817 }       & 0.772        & 0.762        & 0.803        & 0.803        & 0.803        \\
           & $AUC^{pro} , AUC^{pri}$ & \textbf{0.819,} 0.788   & 0.689, 0.642 & \textbf{0.819, 0.814} & 0.757, 0.779 & 0.781, 0.794 & 0.819, 0.788 & 0.819, 0.788 & 0.819, 0.788 \\
           & ACC                     & \textbf{0.759}          & 0.626        & \textbf{0.759 }       & 0.703        & 0.738        & 0.697        & 0.728        & 0.728        \\
           & SPD                     & -0.104         & -0.72        & -0.065       & -0.573       & 0.031        & 0.076        & \textbf{-0.016 }      & 0.071        \\
           & EOD                     & -0.166         & -0.063       & -0.202       & \textbf{0.027}        & -0.283       & -0.191       & -0.205       & -0.246       \\
           & PDR                      & 0.585          & 0.697        & 0.574        & 0.662        & 0.677        & 0.492        & 0.595        & 0.605        \\ \hline
\hline
\end{tabular}
\end{adjustbox}
\end{table*}

We present the results of the performance and fairness metrics for all bias mitigation methods across five datasets in Table~\ref{tab:results}.
Aligning with existing literature, we observe that fairness metrics such as SPD and EOD often yield conflicting results~\citep{kleinberg2016inherent}. It is rarely the same method that returns the best results for both metrics. Furthermore, no single method consistently outperforms all others for one of the metrics. However, DIR stands out for its excellent AUC performance, aligning with its design to preserve rank-ordering within groups~\citep{feldman2015certifying}.

\subsubsection*{Can we use the prediction labels to compare the methods?}
When comparing bias mitigation methods, conventional benchmarking studies often focus on accuracy and fairness~\citep{chen2023comprehensive, zemel2013learning}. However, relying solely on accuracy, which uses prediction labels at a specific threshold, may lead to inappropriate comparisons. As mentioned, there are two reasons why using the prediction labels is not a suitable approach: First of all, the fairness metrics are not yet satisfied, and secondly, different mitigation methods can result in varying positive rates. Both reasons would lead to an additional altering of the prediction labels post-hoc, so comparing them at this stage does not seem sensible. A more comprehensive approach involves assessing the performance of the prediction scores, or comparing the prediction labels when the thresholds have been modified to address both industry constraints and fairness considerations. 

Addressing the first concern, we see that in the large majority of cases, bias mitigation methods fail to satisfy a fairness metric, which is confirmed in literature~\citep{chen2023comprehensive}. Particularly, only postprocessing strategies tend to show a high success rate in achieving the optimized fairness metric, in contrast to the preprocessing and inprocessing methods.
As highlighted in other benchmark studies~\citep{chen2023comprehensive}, we also note that employing bias mitigation methods can even lead to situations that are \textbf{more} unfair in terms of disparities between groups. 

Regarding the second concern, Table~\ref{tab:results} reveals significant positive decision rate disparities among different strategies. This inconsistency poses challenges in real-world applications, where a fixed or reasonably bounded positive decision rate is typically expected~\citep{kwegyir2023repairing}. For example, when attempting to satisfy one of the fairness metrics, practitioners might consider LFR in the Adult Income dataset, despite a slight accuracy loss. However, this choice results in an unexpectedly low positive decision rate (0.5\%), deviating significantly from other methods. This method just predicts almost every instance as negative, which results in a satisfaction of the fairness metrics, but is not realistic in the real-word. Similarly, for the Dutch dataset, MFC leads to the best value for EOD, but it only has a positive decision rate of 20.3\%, while the initial model has a positive decision rate of 41.6\%. DIR leads to the best AUC, but has a positive decision rate of 58.6\%. This demonstrates the second side effect of using bias mitigation methods, namely an altering of the global selection rate.
We argue against treating methods with significantly different positive decision rates as comparable situations. In practice, most real-world applications will have a relatively fixed positive decision rate, and bias mitigation methods must be adapted accordingly. 

\subsubsection*{Evaluate by using the prediction scores}
Both these concerns can effectively be addressed by adjusting the classification threshold(s) of the predictive model. Consequently, it makes more sense to evaluate these mitigation methods based on their prediction scores, rather than on their prediction labels, which are still subject to change. We advocate to compare these mitigation methods based on the AUC score, to assess how well the individuals are ranked within each group, and to generate the prediction labels post-hoc, based on the chosen fairness metric and practical constraints.

It is important not only to examine the overall AUC score across the entire population, but also the AUC scores disaggregated by different groups. Certain bias mitigation methods may yield the optimal ranking for one group but not for another. 
This prompts the need for decision-making: should preference be given to the overall best ranking or to narrowing the gap in rankings between the protected and privileged groups? Alternatively, deploying two separate models could be considered to ensure the best ranking for each subgroup. Given these results, one should evaluate whether a slight improvement in AUC justifies the adoption of distinct models.
Additionally, note that for both the Adult and the Dutch datasets, the LFR method reduces the within-group ranking to an essentially random ordering.
In any scenario, deploying a bias mitigation method that decreases the AUC score for every subgroup seems counterproductive when we want to optimize for both performance and fairness.\footnote{Note that we operate under the assumption that there is no label bias.} Unfortunately, this is a common outcome in practice, as many bias mitigation methods may compromise AUC scores across subgroups in an attempt to adhere to a fairness metric~\citep{mittelstadt2023unfairness}.

Note that the goal of this comparison is not to declare the superior performance of one of the mitigation methods. For this, a more comprehensive benchmark study is needed with a larger number of datasets, machine learning models, and extensive tuning of each bias mitigation method. This was already the goal of multiple other benchmark studies~\citep{chen2023comprehensive,hort2021fairea, reddy2022benchmarking}.
Our primary goal was to show some of the unintended side effects of bias mitigation methods, such as an altering of the global selection rate and the effects within each group, and to emphasize the inadequacy of the current way of benchmarking these methods with each other, based on prediction labels that are still be subject to change.
\section{Discussion}
In this study , we demonstrate two side effects of deploying bias mitigation methods that are not (sufficiently) taken into account.
First, bias mitigation methods not only introduce significant changes between-groups but also within-groups and these changes currently go unnoticed. This reranking process is not necessarily an issue, but deserves more attention. Second, they lead to an altering of the global selection rate which results in very different scenarios. We argue that the current approach to comparing bias mitigation methods is insufficient due to the sole
focus on the output of the prediction model without taking into account the decision-making system~\citep{scantamburlo2024prediction}. 

Regarding the first point, is this reranking process necessarily a bad thing?
We notice in Table~\ref{tab:results} that the bias mitigation methods can result in  slightly better, worse or approximately the same ranking accuracy (measured by the AUC) as the initial machine learning model.
If the ranking improves after using the bias mitigation method, there is no issue, as the method will lead to better rankings that should also be more fair.
But what if the ranking accuracy is significantly worse than the the output of the initial machine learning model? Is this always undesirable?
Not necessarily. Until, and consistent with most literature in fair machine learning, we assumed that we only want to remove bias between groups. However, if we assume there is also bias within each group (so the Affirmative Action Theorem does not hold) and this is undesirable, then it makes sense to use mitigation methods that also shift the ranking within each group, even if they appear worse with respect to the target label.

 If we assume that there is no within-group bias, we propose that for the dual objectives of fairness and accuracy optimization, preprocessing and inprocessing mitigation methods should be adopted only if they improve subgroup rankings. Should they fail to do so, we posit that utilizing the ranking produced by the original machine learning model and then applying Threshold Optimization post-hoc is sufficient and optimal, as it directly addresses the fairness measures without unnecessary within-group reranking. 
 However, this practice is not always possible, as this can lead to situations where two otherwise identical entities are treated differently based solely on a sensitive attribute, a practice that may be unlawful in certain contexts. Furthermore, we can not always assume that the decision-making body has access to the sensitive attribute, and thus is able to do these fairness interventions post-hoc.
 In those settings, it might be better to use preprocessing or inprocessing bias mitigation methods, even if they worsen this trade-off.

Finally, what if the performance of the ranking stays approximately the same, but the ranking itself will be significantly different from the ranking of the initial machine learning model (as measured in Table~\ref{tab:rbo_ranking})? Is this arbitrariness a problem? The opinions on this differ.
The literature on `\textit{predictive multiplicity'}~\citep{marx2020predictive} or `\textit{model multiplicity}'~\citep{black2022model} discusses the situation where there exist many possible models with similar predictive performance but slightly different decisions on individuals, which is comparable to our situation. They argue that multiplicity should be reduced by removing some of the variance that leads to diverging predictions~\citep{cooper2023variance}. 
As we can see, there is no easy answer to this question, and we look forward to more debate about this topic.

\paragraph{Limitations} A first limitation of our study is the assumption of no label bias. This leads us to quite a straddle, which is common in fair machine learning. On the one hand, we presume that our labels are correct to ensure the reliability of our metrics. On the other hand, we recognize that models might need adjustments to correct for bias. This make sense for equal opportunity, where the goal is to ensure that all groups have equal true positive rates, thus requiring a classifier that does not make the situation worse. For demographic parity, the issue is more complex: why would we want to deviate from the labels if they are correct? Nonetheless, adjustments may be necessary to align with external policy or legal requirements.

A second limitation of our study is that it operates under the assumption of having access to a static sensitive attribute, and it will face difficulties in scenarios where these assumptions may not hold. For instance, the assumption of the sensitive attribute being static overlooks the evolving nature of certain attributes, such as gender, where people can identify as other categories over time. Another limitation arises when considering the assumption of unrestricted access to the sensitive attribute. In reality, legal and ethical considerations may impose constraints on obtaining or utilizing certain sensitive information~\citep{haeri2020crucial,veale2017fairer, johnson2021algorithmic,holstein2019improving}. For instance, privacy laws and regulations may restrict the collection or use of specific attributes, further complicating research in fairness.
\bibliographystyle{plainnat}
\bibliography{references}

\newpage
\appendix

\section{Proof of Theorem 3.7} \label{sec:app:proof}
\renewcommand{\theorem}{3.7}
\begin{theorem}\label{teo:condition}
	The Affirmative Action assumption holds if and only if for all $a\in A$ it holds that for almost all $(x,a), (x',a)\in X\times \{a\}$
  \[
    p(Y=1\vert X=x, A=a)\leq p(Y=1\vert X=x', A=a)\Rightarrow S(x,a)\leq S(x',a)
  \]
    \begin{proof}
        $(\Rightarrow)$ Let's suppose by way of contradiction that the implication is not true for some sensitive attribute $a\in A$. Formally this means that there exists a non-zero subset $U\times\{a\}\times U'\times \{a\}\subseteq X\times \{a\}\times X\times \{a\}$ such that for all $x\in U, x'\in U'$ we have 
        \[
          p(Y=1\vert X=x, A=a)\leq p(Y=1\vert X=x', A=a)
          \quad\text{
          and}\quad         
          S(x,a)> S(x',a)
        \]
        Let's now take a threshold $t$ such that 
        \[
        \sup_{x\in U}p(Y=1\vert X=x, A=a)
        \leq t\leq
        \inf_{x'\in U'}p(Y=1\vert X=x', A=a)
        \]
        such threshold $t$ belongs in the interval $(0,1)$ since $p$ is supposed continuous and since $U\times \{a\}$ and $U'\times \{a\}$ have non-zero measure. 
        Let $\hat{Y}$ be a decision associated with the threshold $t$, that is 
        \[
          \hat{Y}(x,a)=\begin{cases}
            1 & \text{if }p(Y=1\vert X=x, A=a)>t\\
            0 & \text{otherwise}
          \end{cases}
        \]
        In particular, almost everywhere we have $\hat{Y}(U\times\{a\})=0$ and $\hat{Y}(U'\times\{a\})=1$. 
		We claim that such decision cannot be optimal according to $S(x,a)$: if it were, it would exist a threshold $t'$ such that almost everywhere
    \[
          \hat{Y}(x,a)=
          \begin{cases}
            1 & \text{if }S(x, a)>t'\\
            0 & \text{otherwise}
          \end{cases}
    \]
    Then for almost all $x\in U$ and $x'\in U'$
		\[
		t'\leq 
		S(x',a)
		<
		S(x,a)
		\leq
		t'
		\]
		which is a contradiction.\\
        \\
        $(\Leftarrow)$ From the hypothesis almost everywhere we have
        \[
          p(Y=1\vert X=x, A=a) = p(Y=1\vert X=x', A=a)\Rightarrow S(x,a)= S(x',a)
        \]
        So people from the same sensitive group with the same fair probabilities are assigned the same score as well. So we can define a bias function $\beta$ such that almost everywhere
        \[
          \beta\left(p(Y=1\vert X=x, A=a), a\right) = S(x,a) 
        \]
        The function $\beta$ takes as input the fair probability and the sensitive attribute and returns the score. By hypothesis, $\beta$ is a non-decreasing function of the fair probability. If we now take a maximal decision $\hat{Y}$ according to $p$, then for all $a\in A$ we can find a decision $\hat{Y}_a\colon X\times\{a\}\to \{0,1\}$ defined as follows
        \[
          \hat{Y}_a(x,a)=
          \begin{cases}
            1 & \text{if }S(x,a)>\beta\left(t, a\right)\\
            0 & \text{otherwise}
          \end{cases} 
        \]
        which are maximal decisions according to $S(x,a)$ and such that $\hat{Y} = \hat{Y}_0\cup \hat{Y}_1$.
    \end{proof}   
\end{theorem}

\section{Correlation plots}
\begin{figure} [ht]
  \centering
    \begin{subfigure}{0.47\linewidth}
    \centering
    \includegraphics[width=\linewidth]{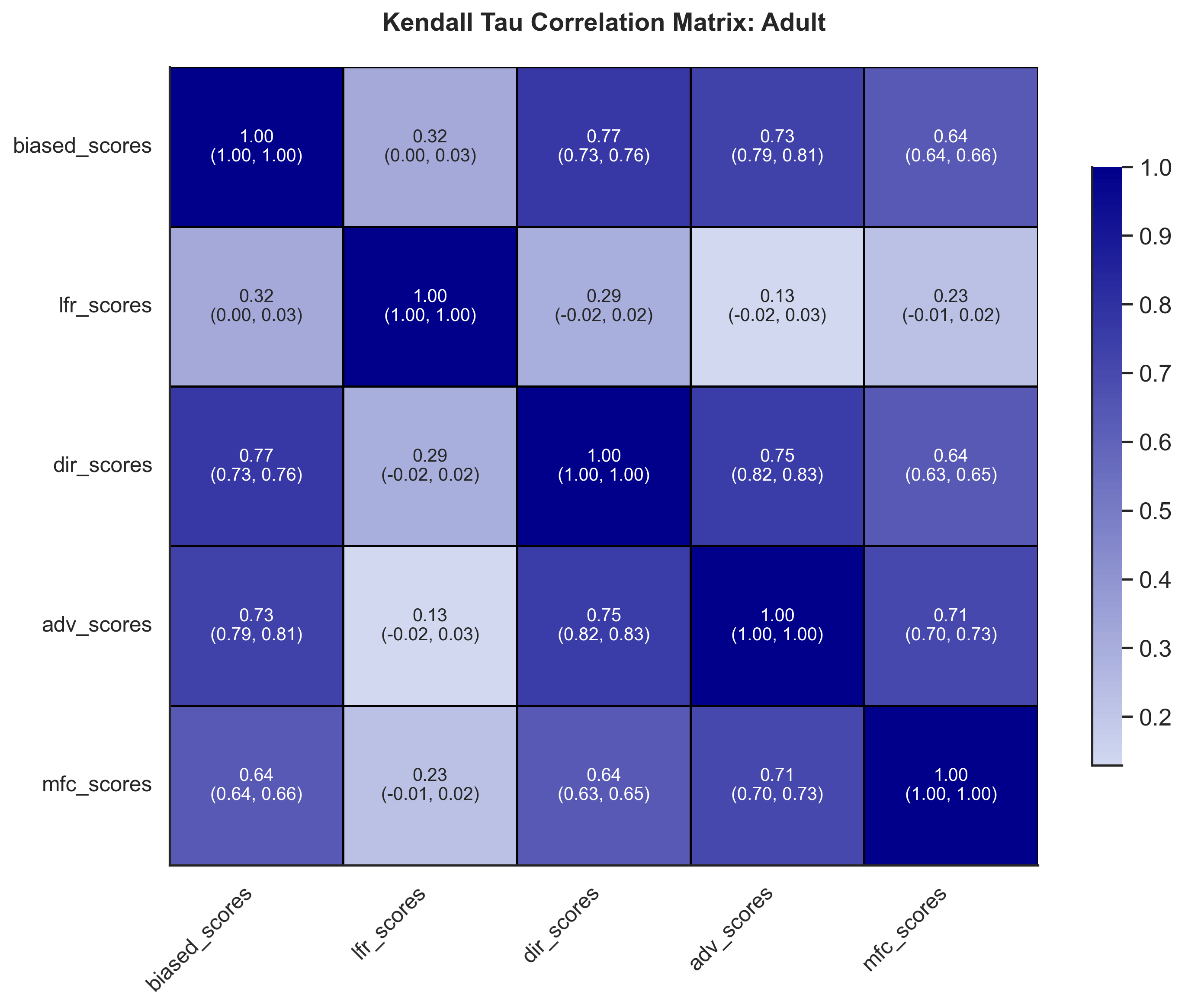}
  \end{subfigure}
  \begin{subfigure}{0.47\linewidth}
    \centering
    \includegraphics[width=\linewidth]{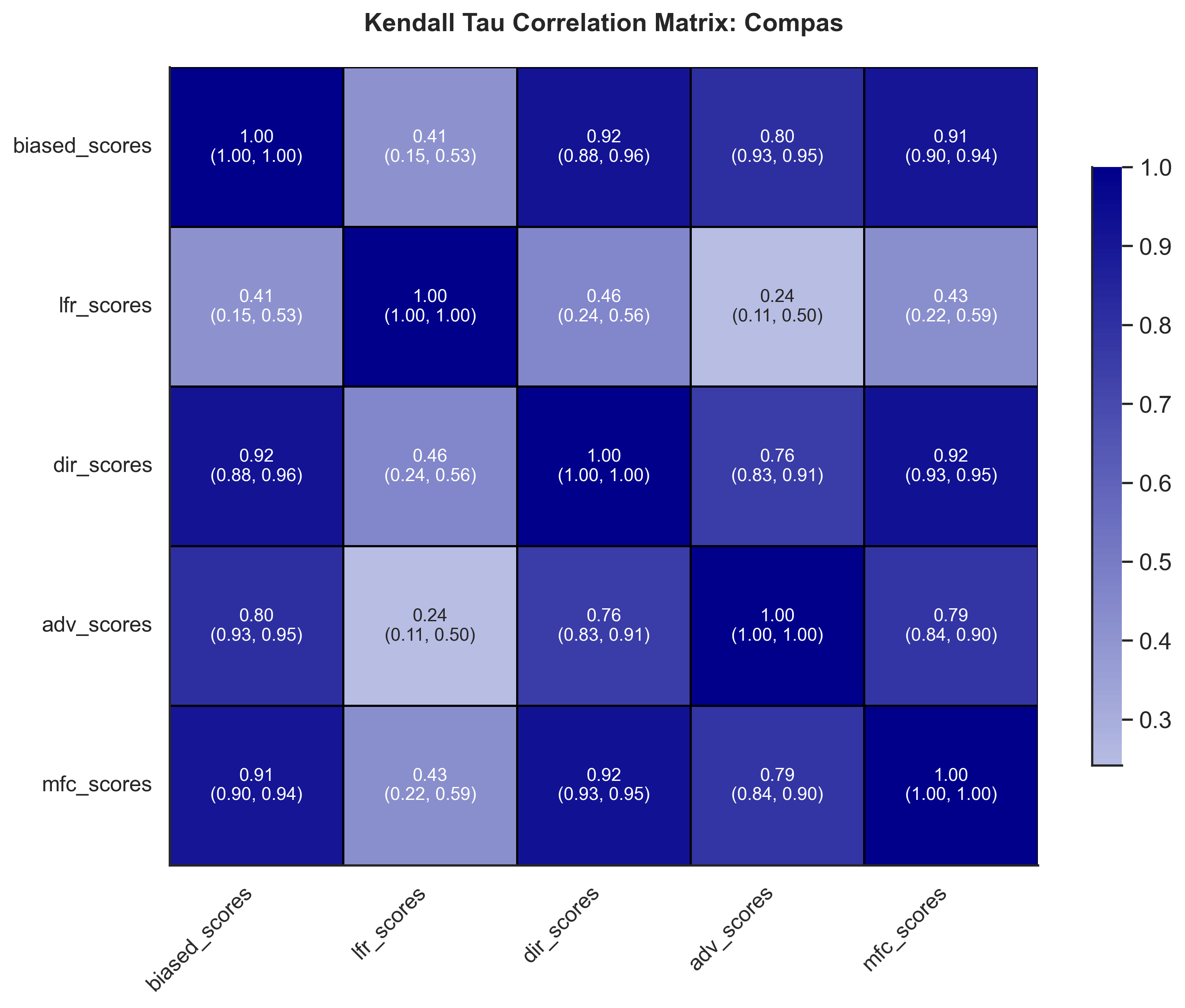}
  \end{subfigure}
    \begin{subfigure}{0.47\linewidth}
    \centering
    \includegraphics[width=\linewidth]{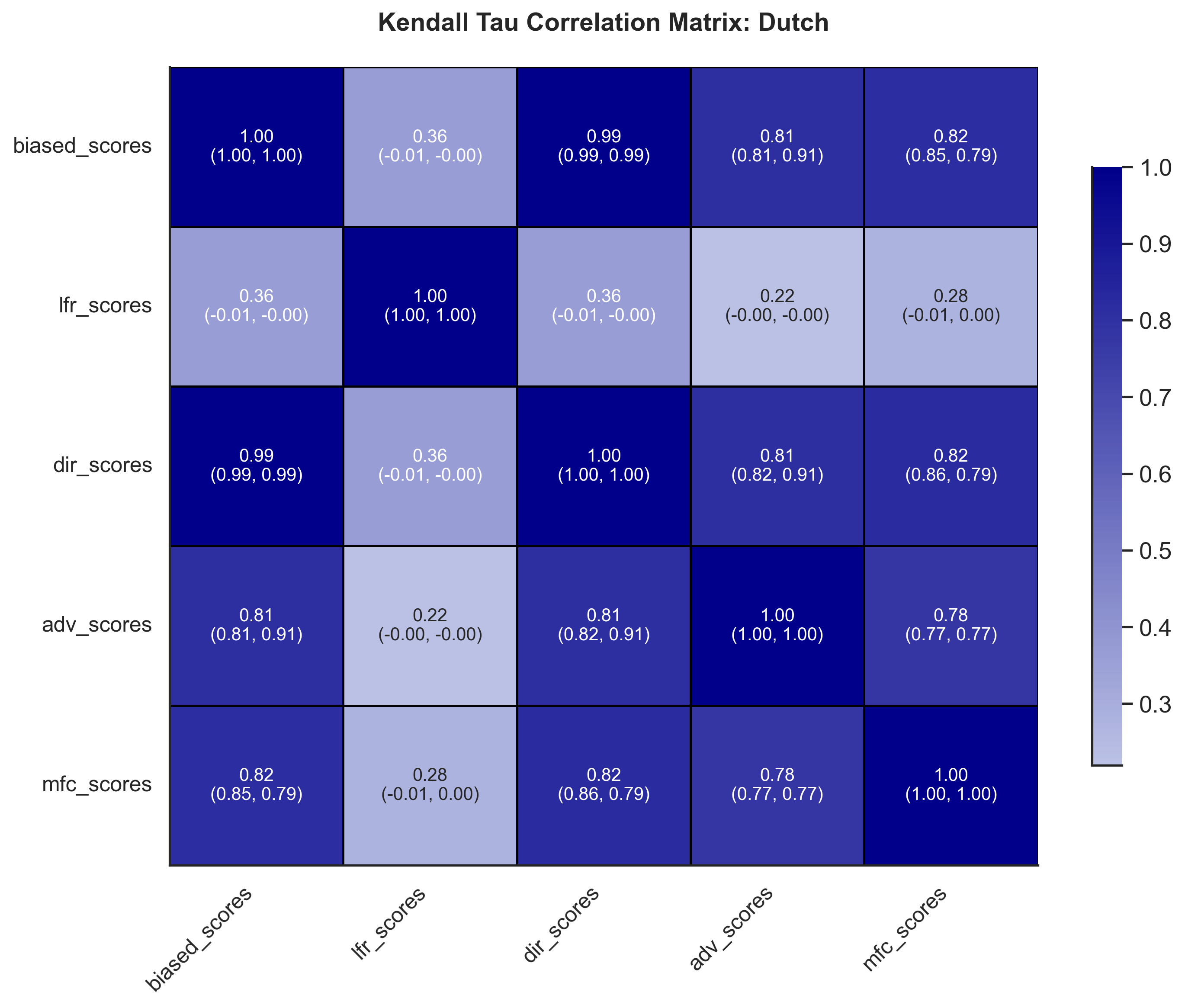}
  \end{subfigure}
    \begin{subfigure}{0.47\linewidth}
    \centering
    \includegraphics[width=\linewidth]{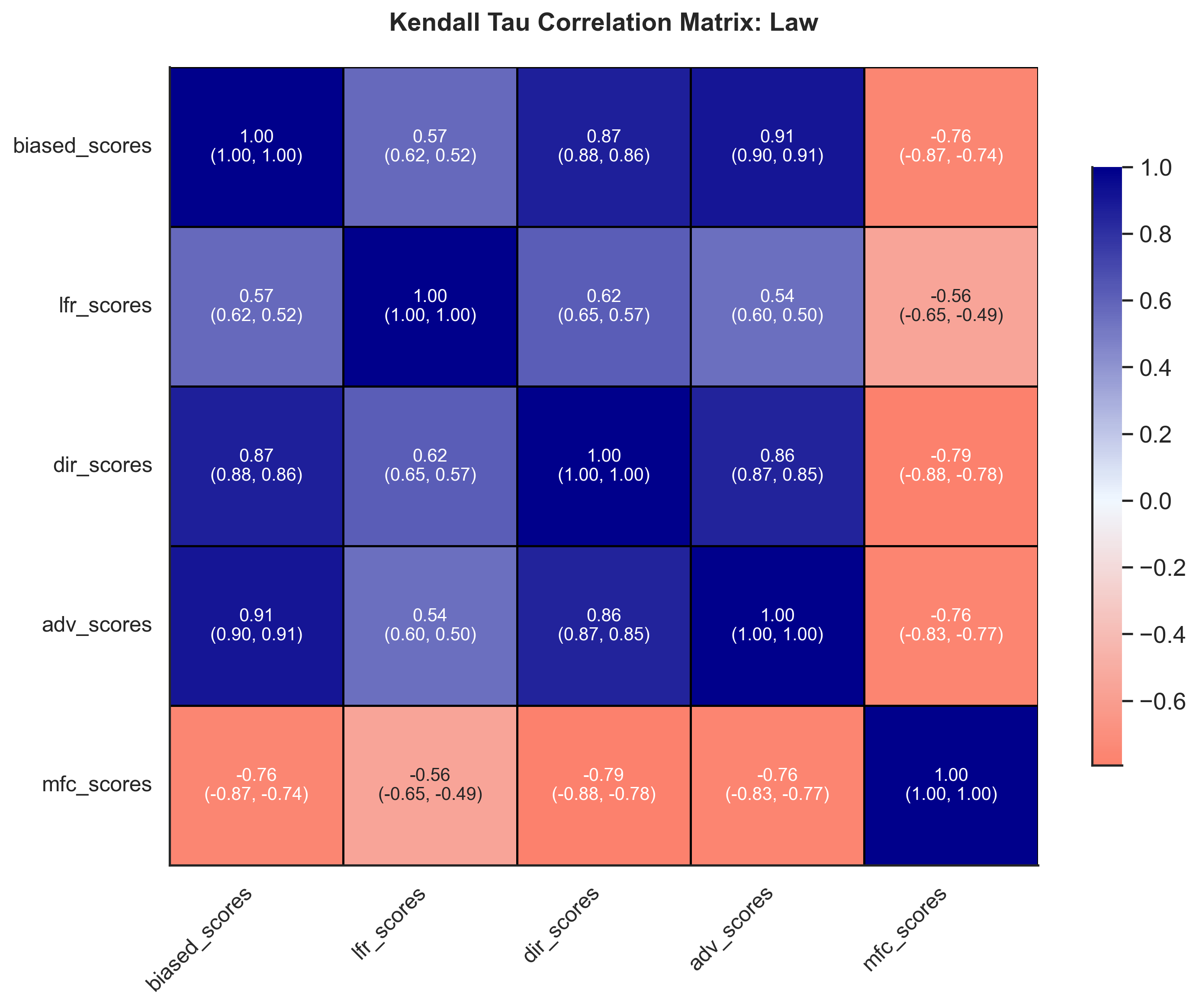}
  \end{subfigure}
  \begin{subfigure}{0.47\linewidth}
    \centering
    \includegraphics[width=\linewidth]{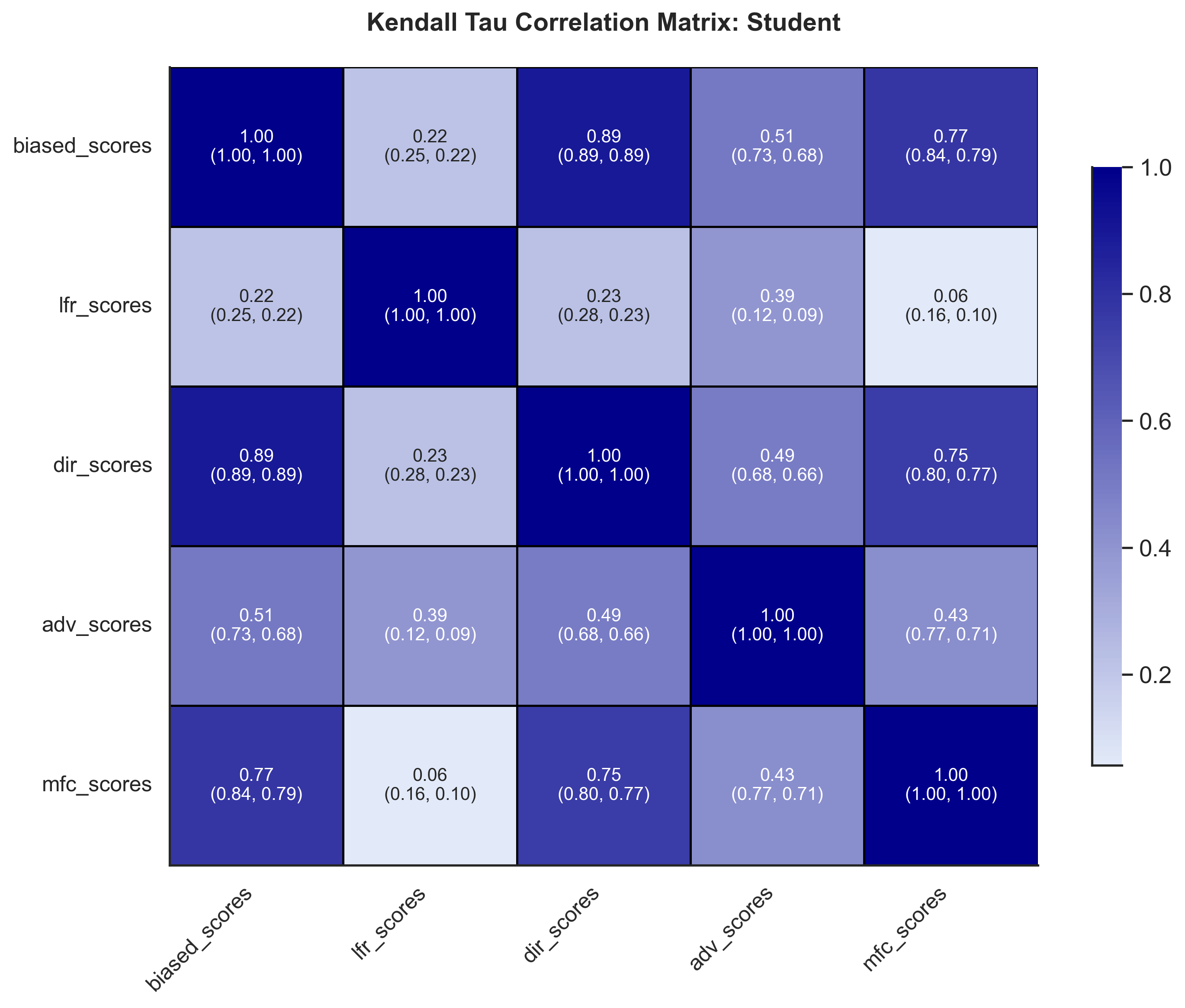}
  \end{subfigure}
    \caption{Correlation plots}
  \label{fig:correlation_plots}
  \end{figure}

\newpage

\section{Score distributions for the other datasets}

\begin{figure} [ht]
  \centering
    \begin{subfigure}{0.32\linewidth}
    \centering
    \includegraphics[width=\linewidth]{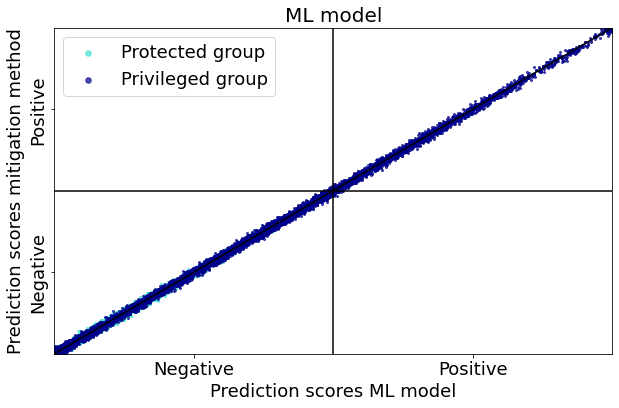}
  \end{subfigure}
  \begin{subfigure}{0.32\linewidth}
    \centering
    \includegraphics[width=\linewidth]{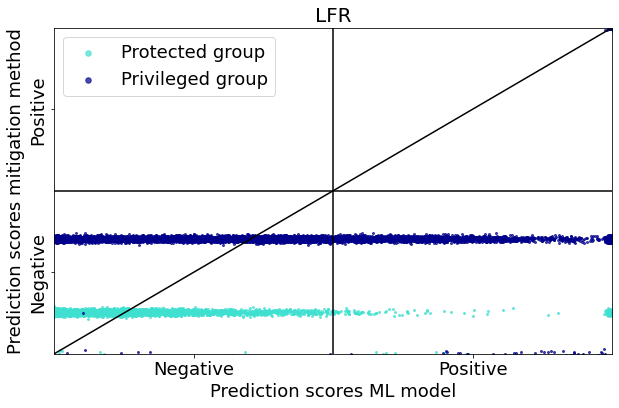}
  \end{subfigure}
    \begin{subfigure}{0.32\linewidth}
    \centering
    \includegraphics[width=\linewidth]{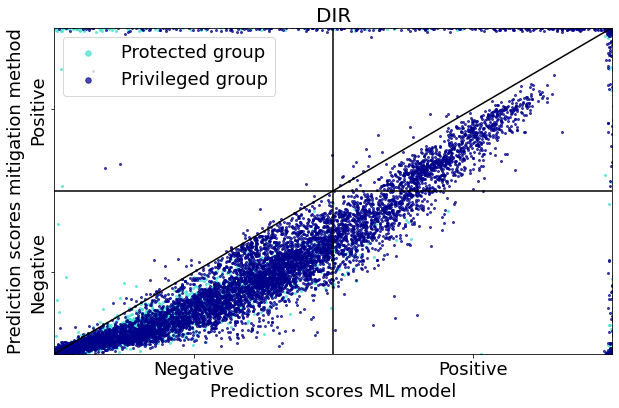}
  \end{subfigure}
    \begin{subfigure}{0.32\linewidth}
    \centering
    \includegraphics[width=\linewidth]{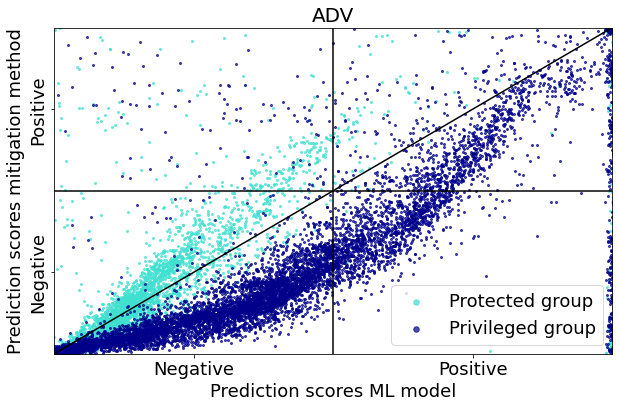}
  \end{subfigure}
  \begin{subfigure}{0.32\linewidth}
    \centering
    \includegraphics[width=\linewidth]{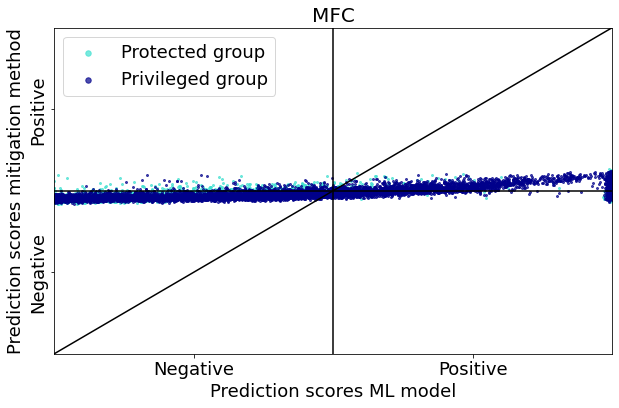}
  \end{subfigure}\\
    \begin{subfigure}{0.32\linewidth}
    \centering
    \includegraphics[width=\linewidth]{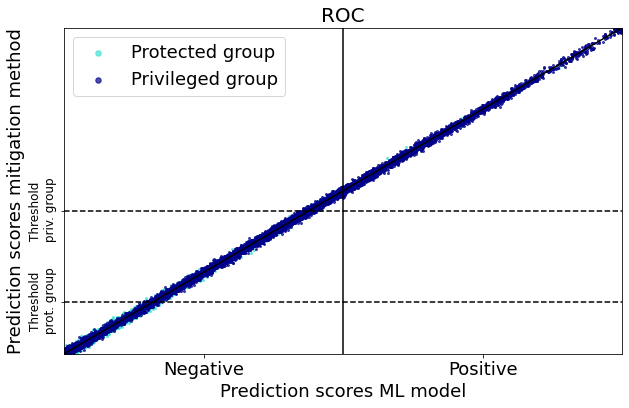}
  \end{subfigure}
      \begin{subfigure}{0.32\linewidth}
    \centering
    \includegraphics[width=\linewidth]{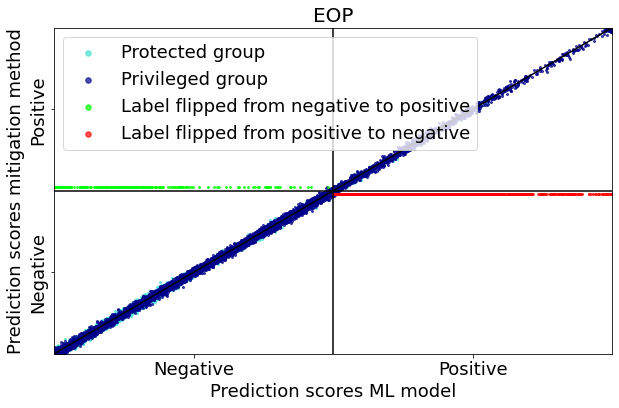}
  \end{subfigure}
  \begin{subfigure}{0.32\linewidth}
    \centering
    \includegraphics[width=\linewidth]{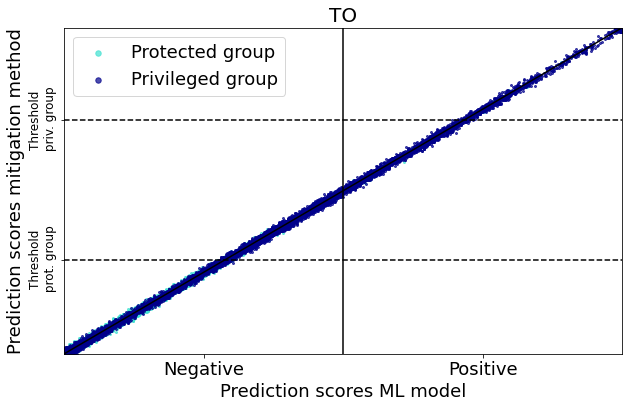}
  \end{subfigure}
  \caption{Score distributions for the Adult dataset}
  \label{fig:scores_adult}
  \end{figure}

  \begin{figure} [ht]
  \centering
    \begin{subfigure}{0.32\linewidth}
    \centering
    \includegraphics[width=\linewidth]{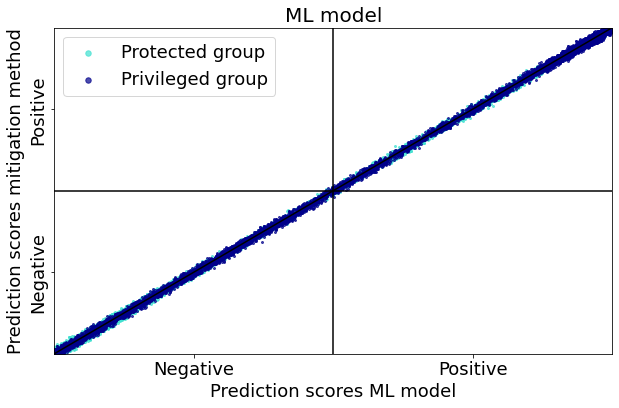}
  \end{subfigure}
  \begin{subfigure}{0.32\linewidth}
    \centering
    \includegraphics[width=\linewidth]{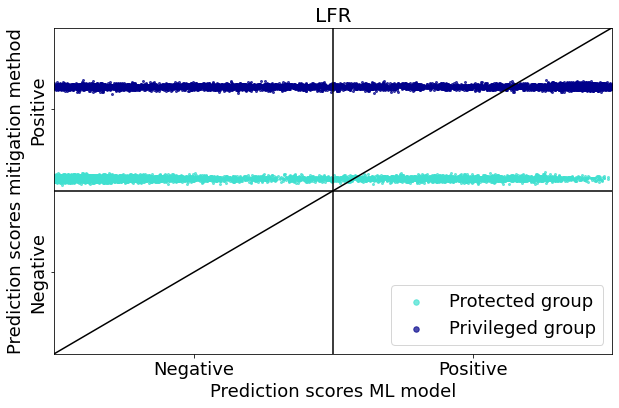}
  \end{subfigure}
    \begin{subfigure}{0.32\linewidth}
    \centering
    \includegraphics[width=\linewidth]{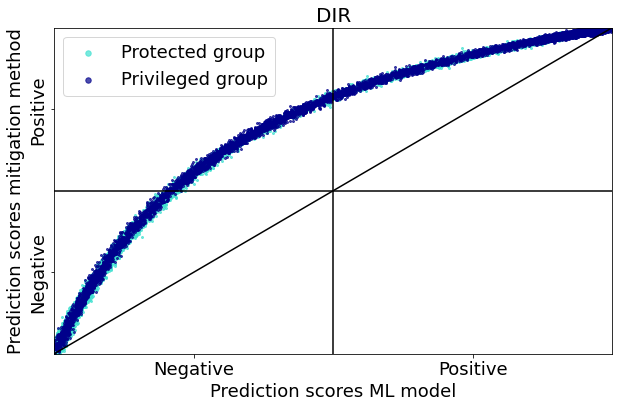}
  \end{subfigure}
    \begin{subfigure}{0.32\linewidth}
    \centering
    \includegraphics[width=\linewidth]{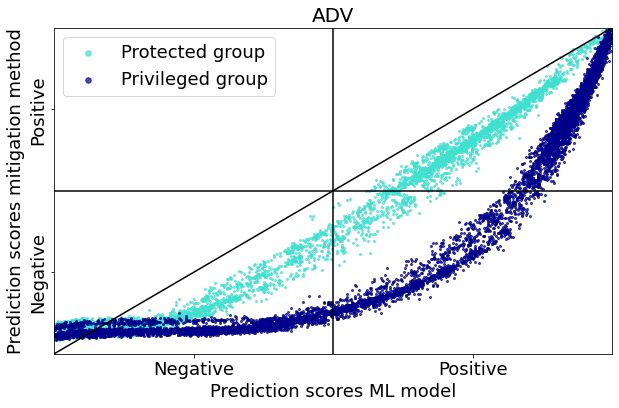}
  \end{subfigure}
  \begin{subfigure}{0.32\linewidth}
    \centering
    \includegraphics[width=\linewidth]{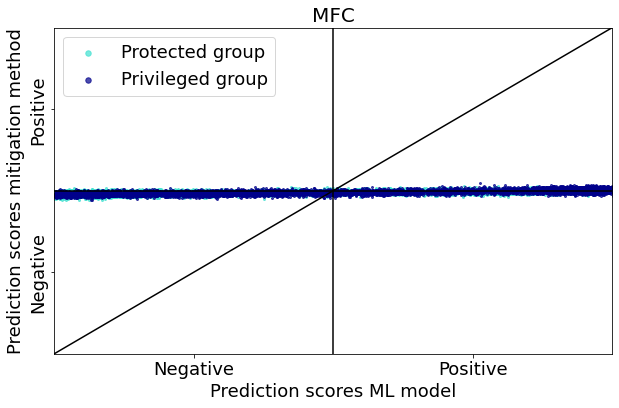}
  \end{subfigure}\\
    \begin{subfigure}{0.32\linewidth}
    \centering
    \includegraphics[width=\linewidth]{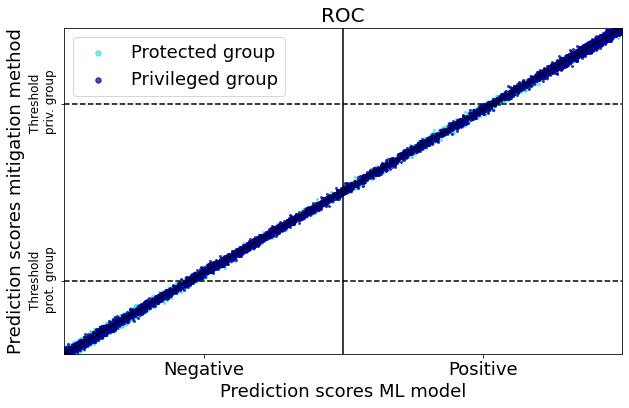}
    \end{subfigure}
    \begin{subfigure}{0.32\linewidth}
    \centering
    \includegraphics[width=\linewidth]{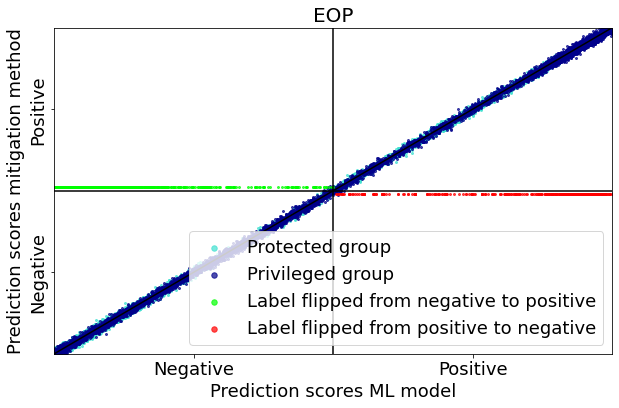}
  \end{subfigure}
  \begin{subfigure}{0.32\linewidth}
    \centering
    \includegraphics[width=\linewidth]{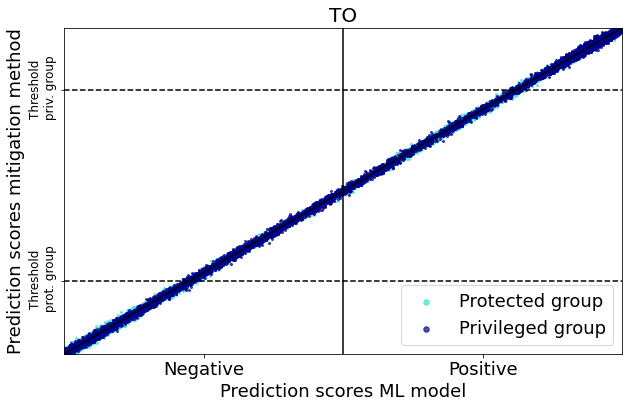}
\end{subfigure}
  \caption{Score distributions for the Dutch dataset}
   \label{fig:scores_dutch}
  \end{figure}

    \begin{figure} [ht]
  \centering
    \begin{subfigure}{0.32\linewidth}
    \centering
    \includegraphics[width=\linewidth]{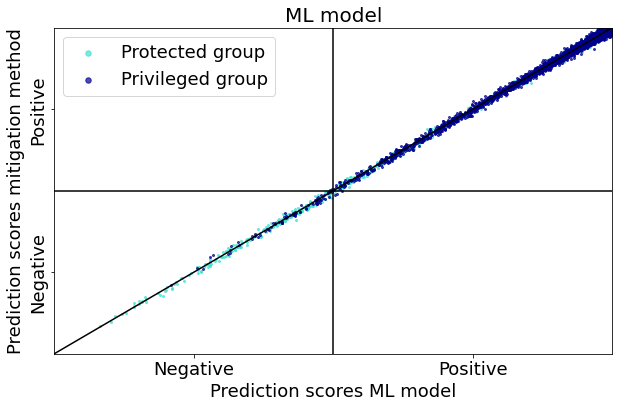}
  \end{subfigure}
  \begin{subfigure}{0.32\linewidth}
    \centering
    \includegraphics[width=\linewidth]{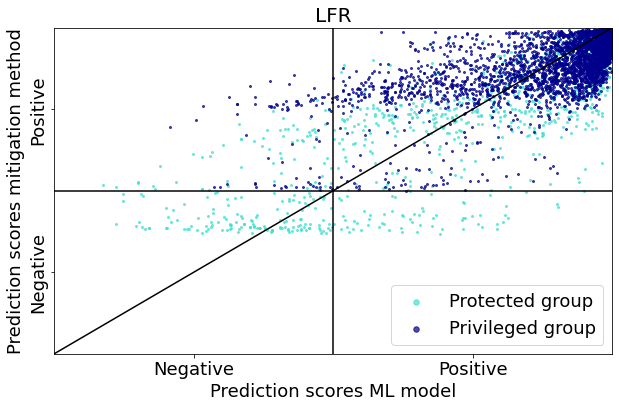}
  \end{subfigure}
    \begin{subfigure}{0.32\linewidth}
    \centering
    \includegraphics[width=\linewidth]{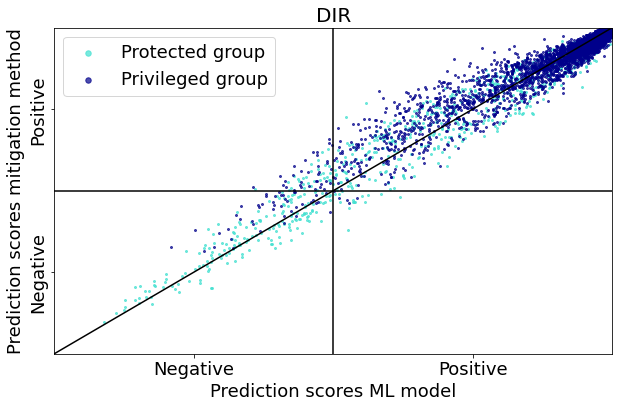}
  \end{subfigure}
    \begin{subfigure}{0.32\linewidth}
    \centering
    \includegraphics[width=\linewidth]{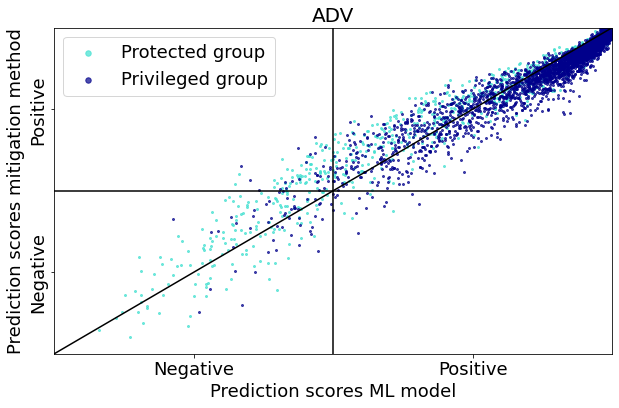}
  \end{subfigure}
  \begin{subfigure}{0.32\linewidth}
    \centering
    \includegraphics[width=\linewidth]{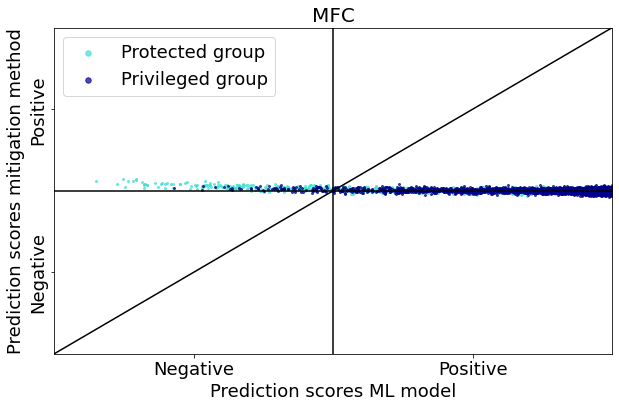}
  \end{subfigure} \\
    \begin{subfigure}{0.32\linewidth}
    \centering
    \includegraphics[width=\linewidth]{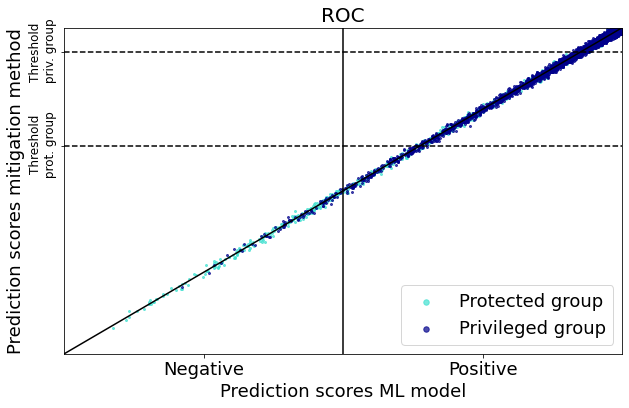}
    \end{subfigure}
    \begin{subfigure}{0.32\linewidth}
    \centering
    \includegraphics[width=\linewidth]{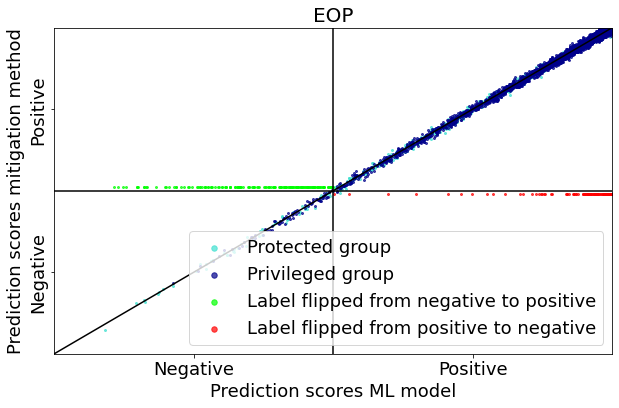}
  \end{subfigure}
  \begin{subfigure}{0.32\linewidth}
    \centering
    \includegraphics[width=\linewidth]{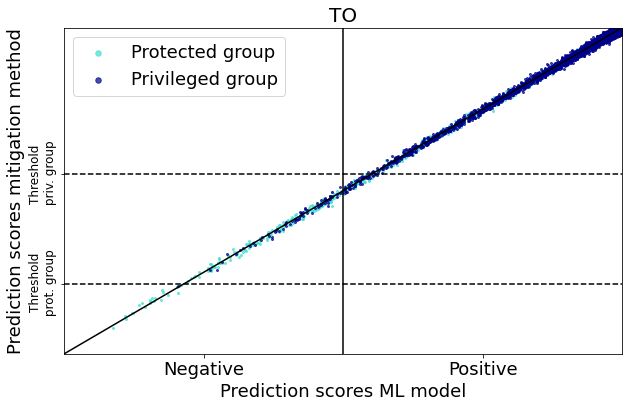}
\end{subfigure}
  \caption{Score distributions for the Law dataset}
   \label{fig:scores_law}
  \end{figure}

    \begin{figure} [ht]
  \centering
    \begin{subfigure}{0.32\linewidth}
    \centering
    \includegraphics[width=\linewidth]{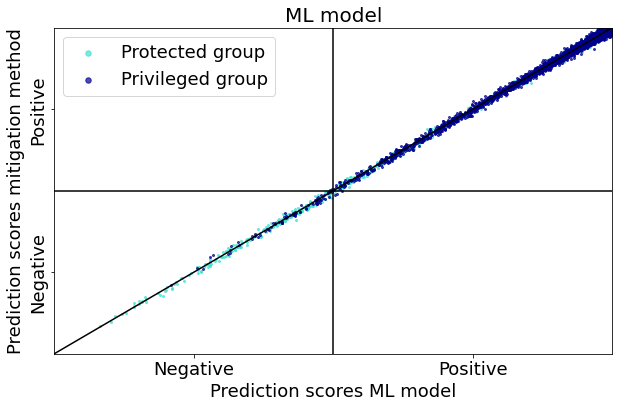}
  \end{subfigure}
  \begin{subfigure}{0.32\linewidth}
    \centering
    \includegraphics[width=\linewidth]{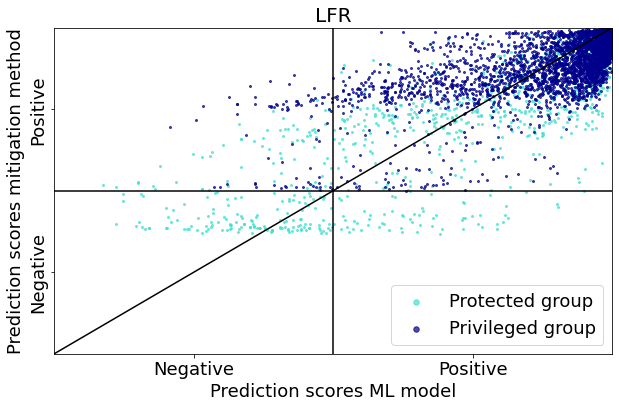}
  \end{subfigure}
    \begin{subfigure}{0.32\linewidth}
    \centering
    \includegraphics[width=\linewidth]{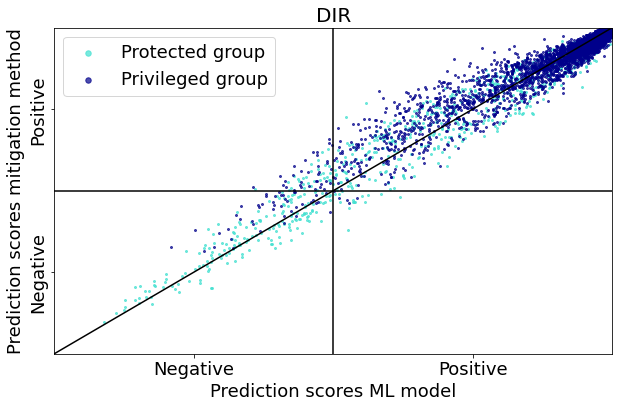}
  \end{subfigure}
    \begin{subfigure}{0.32\linewidth}
    \centering
    \includegraphics[width=\linewidth]{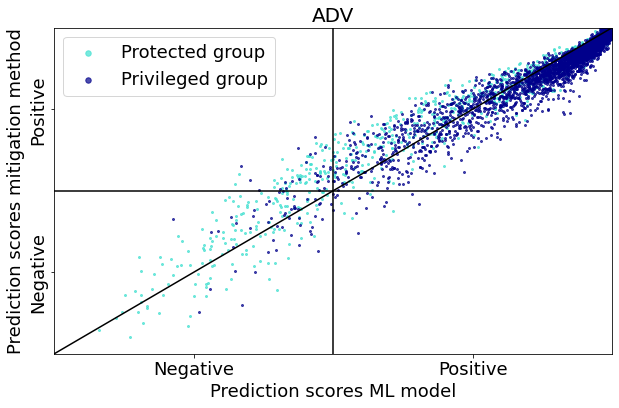}
  \end{subfigure}
  \begin{subfigure}{0.32\linewidth}
    \centering
    \includegraphics[width=\linewidth]{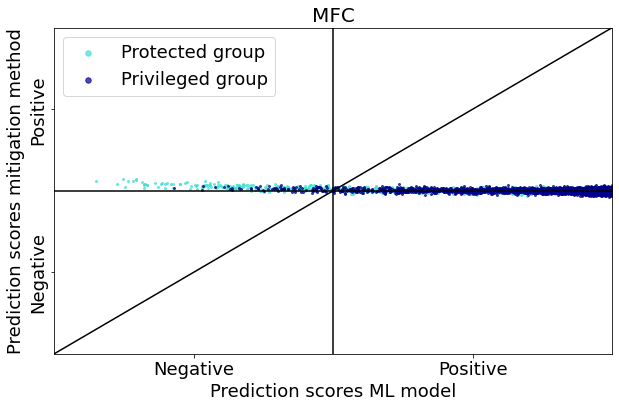}
  \end{subfigure} \\
    \begin{subfigure}{0.32\linewidth}
    \centering
    \includegraphics[width=\linewidth]{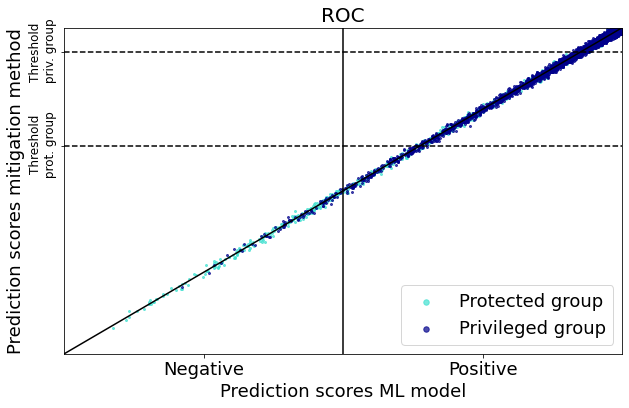}
    \end{subfigure}
    \begin{subfigure}{0.32\linewidth}
    \centering
    \includegraphics[width=\linewidth]{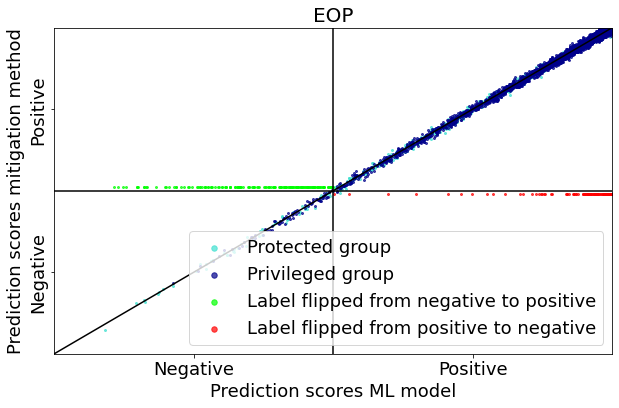}
  \end{subfigure}
  \begin{subfigure}{0.32\linewidth}
    \centering
    \includegraphics[width=\linewidth]{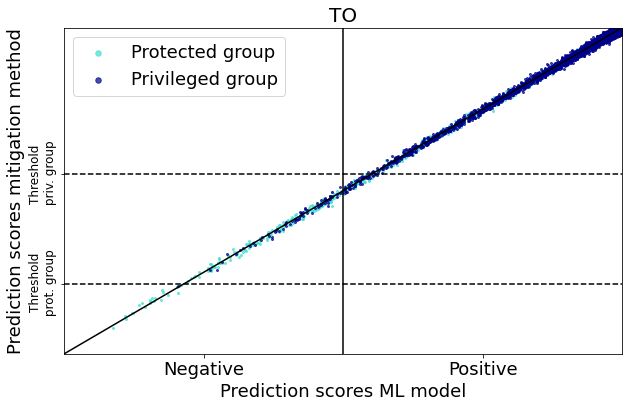}
\end{subfigure}
  \caption{Score distributions for the Student dataset}
   \label{fig:scores_student}
  \end{figure}

\end{document}